\documentclass[10pt,twocolumn,letterpaper]{article}

\usepackage{cvpr}              
\usepackage[dvipsnames]{xcolor}
\usepackage{graphicx}
\usepackage{booktabs}
\usepackage{amsmath}
\usepackage{amssymb}
\usepackage{times}
\usepackage{epsfig}
\usepackage{float}
\usepackage{amsmath}
\usepackage{amssymb}
\usepackage{caption}
\usepackage{subcaption}
\usepackage{adjustbox}
\usepackage{makecell}
\usepackage{bm,multirow}
\usepackage{arydshln}
\usepackage[accsupp]{axessibility}

\long\def\comment#1{} 

\newcommand{\beq}{\begin{equation}}
\newcommand{\eeq}{\end{equation}}
\newcommand{\beqa}{\begin{eqnarray}}
\newcommand{\eeqa}{\end{eqnarray}}

\usepackage[pagebackref,breaklinks,colorlinks]{hyperref}

\usepackage[capitalize]{cleveref}
\crefname{section}{Sec.}{Secs.}
\Crefname{section}{Section}{Sections}
\Crefname{table}{Table}{Tables}
\crefname{table}{Tab.}{Tabs.}

\def\cvprPaperID{11667} 
\def\confName{CVPR}
\def\confYear{2022}
\begin{document}

\title{CLIPstyler: Image Style Transfer with a Single Text Condition }
\author{ Gihyun Kwon$^{1}$ \quad\quad Jong Chul Ye$^{1,2}$\\
Dept. of Bio and Brain Engineering$^1$, 
Kim Jaechul Graduate School of AI$^2$, KAIST
\\
\tt\small \{cyclomon,jong.ye\}@kaist.ac.kr
}

\twocolumn[{%
\renewcommand\twocolumn[1][]{#1}%
\maketitle
\begin{center}
\centering
\includegraphics[width=0.97\linewidth]{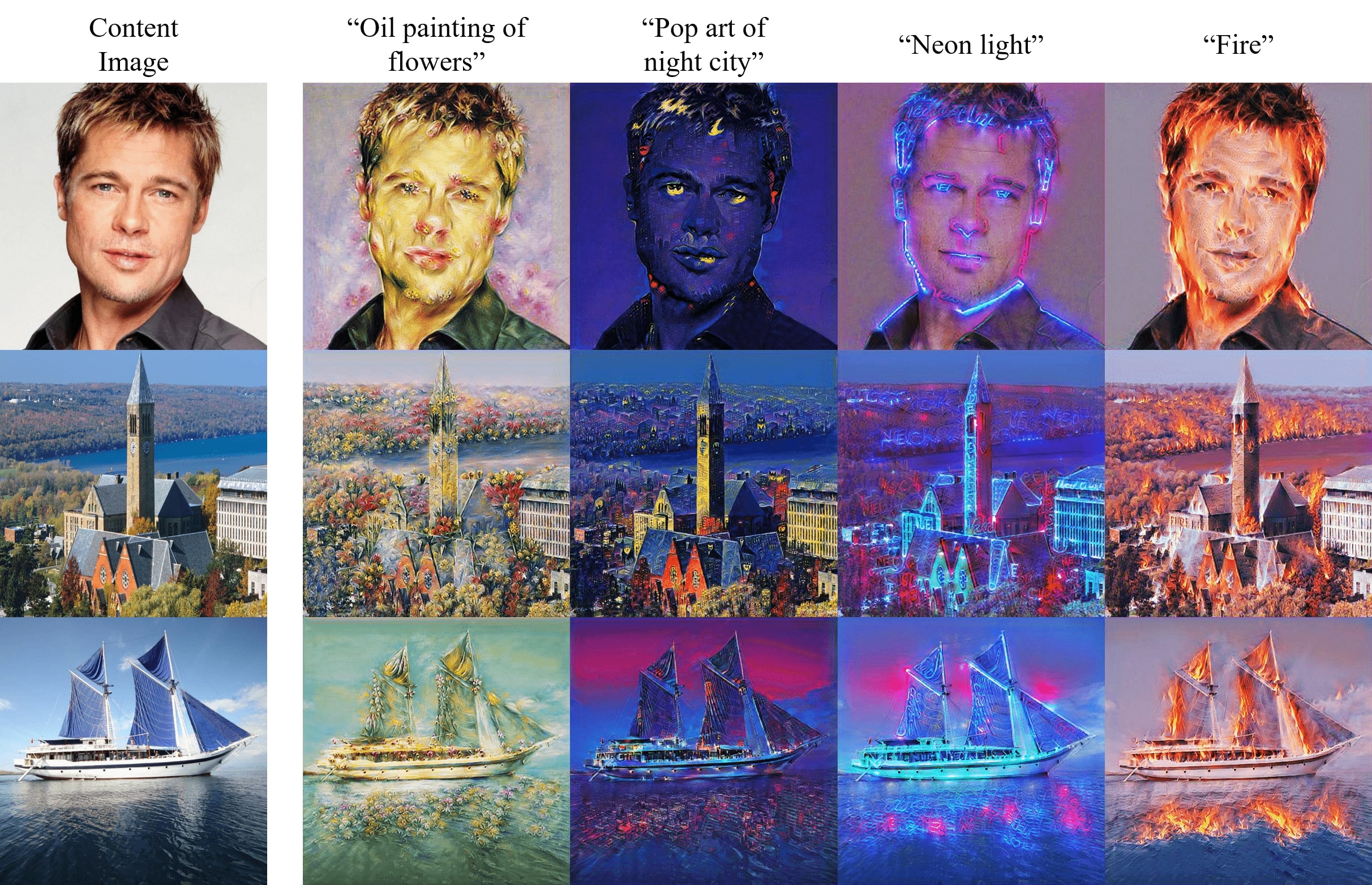}
 \vspace*{-0.2cm}
\captionof{figure}{Our style transfer results on various text conditions. Translated images have spatial structure of the content images with realistic textures corresponding to the text.}
\label{fig:first}
\end{center}}]
\begin{abstract}
Existing neural style transfer methods require reference style images to transfer texture information of  style images to content images.
However, in many practical situations, users may not have reference style images but still be interested in transferring styles by just imagining them.
In order to deal with such applications, we propose a new framework that enables a style transfer `without' a style image, but only with a text description of the desired style.  
Using the pre-trained text-image embedding model of CLIP, we demonstrate the modulation of the style of content images only with a single text condition. 
Specifically, we propose a  patch-wise text-image matching loss with multiview augmentations for realistic texture transfer.
 Extensive experimental results confirmed the successful  image style transfer with realistic textures that reflect semantic query texts.
\end{abstract}

\section{Introduction}

Style transfer aims to transform a content image by transferring the semantic texture of a style image. The seminar work of neural style transfer proposed by Gatys {\it{et al.}}\cite{gatys} uses a pre-trained VGG network to transfer the style texture by calculating  the style loss that matches the Gram matrices of the content and style features.
Their style loss has become a standard in later works including stylization through pixel optimization for a single content image\cite{elad}, arbitrary style transfer which operates in real-time for various style images\cite{adain,li1,yoo,liu}, and optimizing feedforward network for stylizing each image\cite{johnson,ulyanov}. 

Although these approaches for style transfer can successfully create visually pleasing new artworks by transferring styles of famous artworks to common images,
they require a reference style image to change the texture of the content image. 
However, in many practical applications, reference style images are not available to users, but the users are still interested in `imitiating' the texture of the style images.
For example, users can imagine being able to convert their own photos into Monet or Van Gogh styles without ever owning paintings by the famous painters.
Or you can convert your daylight images into night images by mere imagination.
In fact,  to overcome this limitation of the existing style transfer and create a truly creative artwork, we should be able to transfer a completely novel style that we imagine.

Toward this goal, several methods have attempted to manipulate images with a text condition which conveys the desired style. Using pre-trained text-image embedding models, 
these method usually deliver semantic information of text condition  to the visual domain. However, these methods often have disadvantages in that semantics are not properly reflected due to the performance limitations of the embedding model\cite{zhang1,zhang2}, and the manipulation
is restricted to a specific content domain (such as human face) as the method heavily rely on pre-trained generative models\cite{styleclip}.

To address this, here we propose a novel image style transfer method to deliver the semantic textures of text conditions using recently proposed text-image embedding model of CLIP\cite{clip}. 
Specifically, rather than resorting to pixel-optimization or manipulating the instance normalization layer as in AdaIN \cite{adain},
we propose to train a lightweight CNN network that can express the texture information with respect to text conditions and produce realistic and colorful results. 
More specifically,  the content image is transformed by the lightweight CNN to follow the text condition by matching the similarity 
between the CLIP model output of transferred image and the text condition. 
Furthermore, when the network is trained for multiple content images,
our method enables text-driven style transfer regardless of content images.

Our method comes from several technical innovations in the implementation.
First,  instead of optimizing the loss by using the image directly, we propose to use a patch-wise CLIP loss to guide the network to function as a brushstroke. 
Specifically, to calculate the proposed loss, we first sample patches of the output image and apply augmentation with different perspective views. 
Afterwards, we obtain the CLIP loss by calculating the similarity between the query text condition and the processed patches. By applying this patch-wise CLIP loss, 
we found that we can transfer styles to each local area of the content image. Furthermore, the augmentation induces the patch style to be more vivid and diverse.
Additionally, 
to overcome the patch-dependent over-stylization problem, 
we propose a novel threshold regularization so that the patches with abnormally high scores do not affect the network training.

Extensive experimental  results show that our model can transfer a variety of unique styles based on text conditions, which enable a wider range of style transfer than the existing methods using style images.


\section{Related works}
\subsection{Style Transfer}

Inspired by Gatys et al. \cite{gatys} who proposed an iterative pixel-optimization 
 by jointly minimizing content and style
losses,
 Johnson et al.\cite{johnson} and Ulyanov et al.\cite{ulyanov} proposed to train a stylization feed-forward network with using the same loss function by Gatys et al. \cite{gatys}.
By extending the aforementioned single-content style transfer approaches, 
Li et al.\cite{li1,li2} proposed the whitening and coloring transform (WCT) method to transform the content features to follow the statistic of style features.
Huang et al.\cite{adain} proposed Adaptive Instance Normalization (AdaIN) in which the mean and standard deviation of the style image features are applied to normalized feature statistics of content images. Li et al.\cite{li3} proposed a linear transformation between content and style features for fast style transfer on images and videos. 

Recently, Yoo et al.\cite{yoo} proposed a wavelet transform based WCT for better preservation of content information for photo-realistic style transfer. Park et al.\cite{park} proposed a style attentional  network (SANet) so that the style can refer the content feature information. Svoboda et al.\cite{svoboda} proposed style transfer with graph convolutional network to combine style and content in latent spaces. Deng et al.\cite{deng} suggested multiple adaptation module which use spatial attention as content feature and channel attention as style features. 

More recently, 
Liu et al.\cite{liu} proposed adaptive attention normalization as a improved attention-based style transfer of SANet\cite{park}. Xu et al.\cite{xu} proposed a new framework of dynamic residual block to integrate style and content features of generative models. 
Hong et al.\cite{hong} introduce a domain-aware style transfer method in which the model transfer the domain-aware information along with style. Kotovenko et al.\cite{kotovenko} focused on the brushstroke property of Bezier curves, and proposed to optimize parameters of modelled quadratic Bezier curves instead of pixels.

Although these methods have shown successful results, the methods require style images in order to make the content image to follow the texture of the target style. 

\subsection{Text-guided synthesis}

In the existing text-guided image synthesis, 
the encoders for text embedding work as guide conditions for generative models. Zhang et al.\cite{zhang1,zhang2} integrated text conditions to multi-scale generative model for high-quality image synthesis. AttnGAN\cite{attngan} further improved the performance with attention mechanism on text and image features. ManiGAN\cite{manigan} proposed a modules for simultaneously embedding the text and image features. 

Recently, OpenAI introduced CLIP\cite{clip} which is a high-performance text-image embedding models trained on 400M text-image pairs. CLIP model have shown a state-of-the-art performance on connecting text and image domain. With powerful representation embedding of CLIP, there are several approaches which can manipulate the image with text conditions. StyleCLIP\cite{styleclip} performed attribute manipulation with exploring learned latent space of StyleGAN\cite{stylegan}. They successfully controlled the generation process  by finding an appropriate vector direction towards the given text condition. However, StyleCLIP has limitation as the latent exploration can manipulate images within the trained domain. Therefore, StyleGAN-NADA\cite{nada} proposed a model modification method with using text condition only, and modulates the trained model into a novel domain without additional training images.

Although these models could manipulate the images with text conditions, they are heavily dependent on pre-trained generative models. Therefore, the generated images are confined to trained image domains.  In our model, we can transfer the texture of text conditions to the source image regardless of the domain of images, which was not considered in the aforementioned generative model based manipulations.

%

\section{Method}
\subsection{Basic framework of CLIPstyler}

As emphasized before, the purpose of our framework is to transfer the semantic style of target text $t_{sty}$ to the content image $I_c$ through the pre-trained text-image embedding model CLIP \cite{clip}. The major difference from the existing methods is that in our model, there is no style image $I_s$ to use as a reference.

Since our model aims to obtain semantically transformed images with a sole supervision from CLIP, we have several technical issues to solve: 1) how to extract the semantic `texture' information from CLIP model and apply the texture to the content image, and 2) how to regularize the training so that the output image is not qualitatively impaired. 

\begin{figure}[!t]
\centering
\vspace*{-0.5cm}
\includegraphics[width=1.0\linewidth]{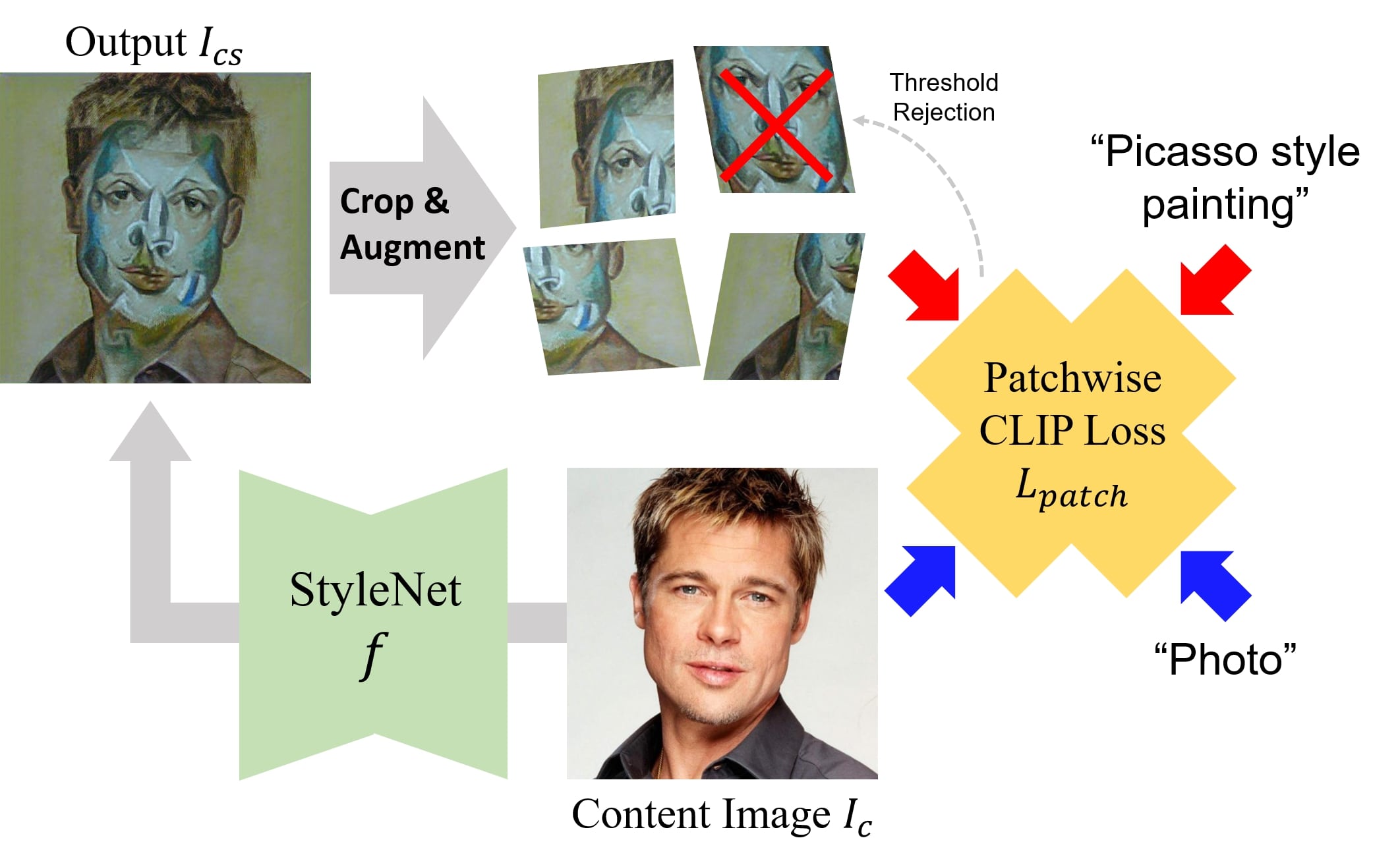}
\vspace*{-0.5cm}
\caption{Overall schematics of our proposed patch-wise CLIP loss. We optimize the neural network  $f$ using the loss functions. 
}
\label{fig:method}
\end{figure}

The specified architecture of our method is shown in Fig.~\ref{fig:method}. When a content image $I_c$ is given, we aim to obtain the style transfer output $I_{cs}$.  Unfortunately, we experimentally found that the desired texture is not reflected when using traditional pixel optimization method.
To solve the problem, we introduced a CNN encoder-decoder model $f$ that can capture the hierarchical visual features of the content image and simultaneously stylize the image in deep feature space to obtain a realistic texture representation. Therefore, our stylized image $I_{cs}$ is $f(I_c)$, and our final goal is to optimize the parameter of $f$ so that it can make the output to have target texture. 

\subsection{Loss function}

\noindent \textbf{CLIP loss:} To guide the content image to follow the semantic of target text, the simplest CLIP-based image manipulation approach\cite{styleclip} is to minimize the global clip loss function which is formulated as:
\begin{equation}
     L_{global} = D_{CLIP}(f(I_c),t_{sty}),
\end{equation}
where $D_{CLIP}$ is the CLIP-space cosine distance. This loss function transforms the output image of the whole frame to follow the semantic of the textual condition. However, when such  global CLIP loss is used, often the output quality is corrupted, and the stability is low in the optimization process. 

To solve the problem, StyleGAN-NADA\cite{nada} proposed a directional CLIP loss that aligns the CLIP-space direction between the text-image pairs of source and output. So, we also employ the directional CLIP loss, which can be defined in our case as:
\begin{align}
\Delta T = E_{T}(t_{sty}) - E_{T}(t_{src}),\nonumber\\
\Delta I = E_{I}(f(I_c)) - E_{I}(I_c),\nonumber\\
L_{dir} = 1-\frac{\Delta I\cdot \Delta T}{|\Delta I||\Delta T|},
\end{align}
where $E_I$ and $E_T$ are the image and text encoders of CLIP, respectively;  and $t_{sty}$, $t_{src}$ are the semantic text of the style target and the input content, respectively. When we use natural images for content, $t_{src}$ is simply set as ``Photo".

\noindent \textbf{PatchCLIP loss:} Although the proposed $L_{dir}$ shows good performance in modifying the pre-trained generative model, it does not perfectly match to ours because our goal is to apply the semantic texture of $t_{sty}$ to a given content image. It is also shown in our result that solely using $L_{dir}$ in our framework decreased the quality of the output.

To overcome the shortcomings of existing CLIP losses, we propose a novel PatchCLIP loss for texture transfer. Inspired by the idea of the original style loss in Gatys et al. \cite{gatys}, which is to deliver the spatially invariant information, we found that a similar effect can be obtained by minimizing CLIP loss functions with respect to group of patches that are extracted from $I_{cs}$ in arbitrary locations.

Specifically, we randomly crop sufficient number of patches from $I_{cs}$. At this stage, the size of the cropped image is fixed. For all of $N$ cropped patches $\hat{I}^N_{cs}$, we apply random geometrical augmentation to the cropped patches before calculating the CLIP directional loss. {Inspired by Frans et al.\cite{clipdraw}}, we found that using augmentations on each patch assist the network to represent more vivid and diverse textures.  

Although there are many possible types of augmentations, we propose to use perspective augmentation. 
With using perspective augmentation,  all patches are guided to have the same semantic when viewed in multiple points so that
 the semantic information of the CLIP model can be reconstructed as more 3D-like structures. 

\noindent \textbf{Threshold rejection:} Due to the stochastic randomness of patch sampling and augmentations, our method often suffer from over-stylization in which the style network $f$ is optimized on specific patches that are easy to minimize the loss scores. To alleviate the problem, we include regularization to reject the gradient optimization process for high-scored patches. With a given threshold value $\tau$, we simply nullify the calculated loss of corresponding patches. Therefore, our proposed patch-wise CLIP loss is defined as:
\begin{align}
\Delta T = E_{T}(t_{sty}) - E_{T}(t_{src}),\nonumber\\ 
\Delta I = E_{I}(aug(\hat{I}^i_{cs})) - E_{I}(I_c),\nonumber\\
l^i_{patch} =1-\frac{\Delta I\cdot \Delta T}{|\Delta I||\Delta T|},\nonumber \\
L_{patch} = \frac{1}{N} \sum^N_i R(l^i_{patch},\tau) \\
\text{where} \quad R(s,\tau)= 
\begin{cases} 
0, \quad\text{if} \quad s\leq \tau \nonumber\\
s , \quad \text{otherwise}
\end{cases}
\end{align}
where $\hat{I}^i_{cs}$ is the $i$-th cropped patch from the output image, $aug$ is random perspective augmentation, and $R(\cdot,\cdot)$ represents the threshold function.

\noindent \textbf{Total loss:} For overall loss function, we use four different losses. First, we use standard directional CLIP loss $L_{dir}$ to modulate the whole part of the content image (e.g. color tone, global semantics). Second, we add our proposed PatchCLIP loss $L_{patch}$ for local texture stylization. On top of CLIP loss functions, to maintain the content information of input image, we include the content loss $L_c$ with calculating the mean-square error between the features of content and output images extracted from the pre-trained VGG-19 networks similar to the existing work by Gatys et al. \cite{gatys}. Finally, to alleviate the side artifacts from irregular pixels, we include the  total variation regularization loss $L_{tv}$. Therefore, our total loss function is formulated as:
\begin{align}
    L_{total} = \lambda_d L_{dir}+ \lambda_p L_{patch} + \lambda_c L_c + \lambda_{tv} L_{tv}
\end{align}

\section{Results}

\begin{figure*}[t]
\vspace*{-0.5cm}
\centering
\includegraphics[width=1.0\linewidth]{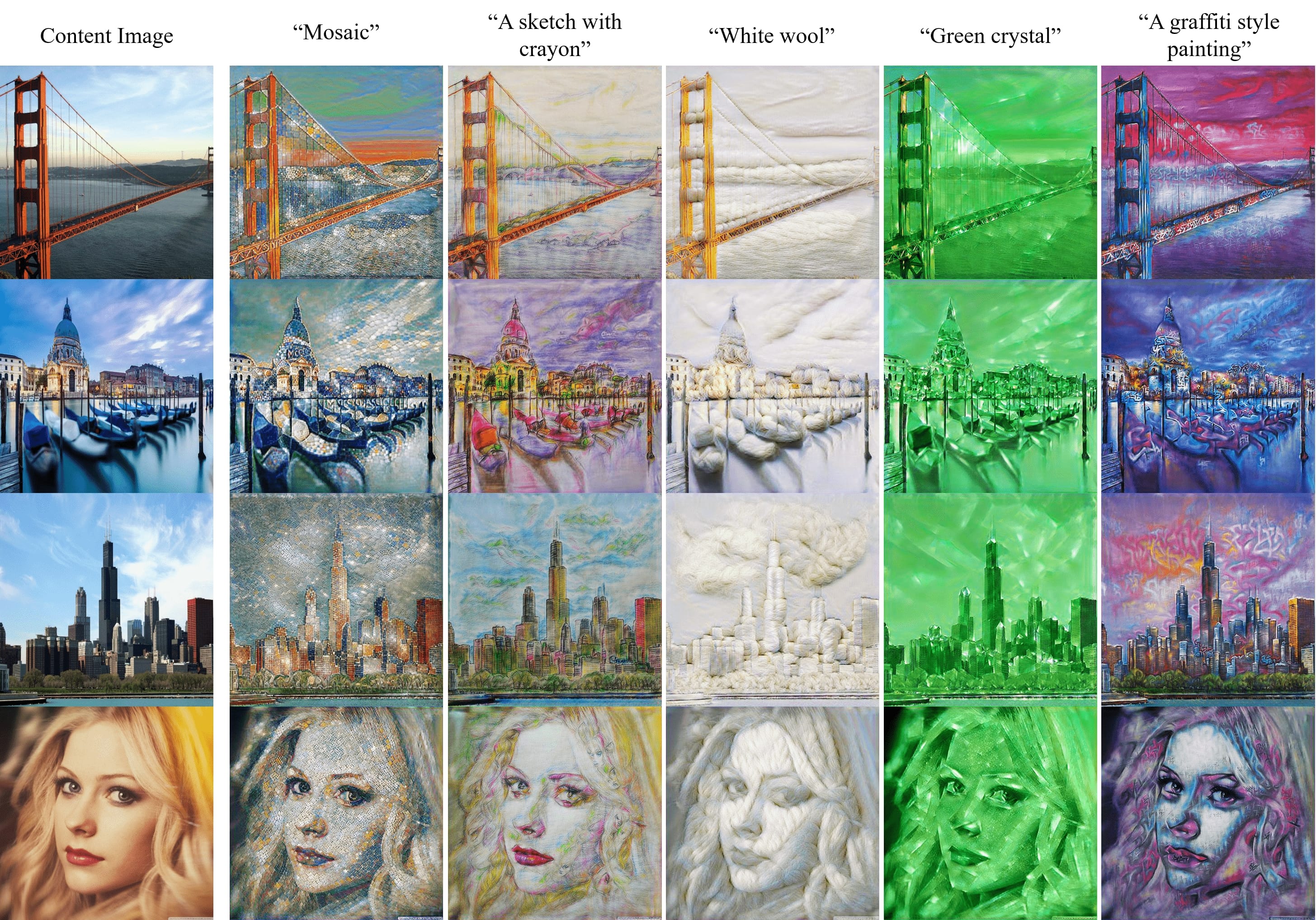}
\vspace*{-0.5cm}
\caption{Style transfer results on various query text conditions. Our method can synthesize realistic textures which reflect the text conditions.  Additional results are in our Supplementary Materials.
}
\label{fig:ours}
\end{figure*}

\subsection{Experiment settings}
Our content image can have any resolution sizes, but considering the resource capacity, we use 512$\times$512 resolution for all content images. For training, we set $\lambda_d$, $\lambda_p$, $\lambda_c$, and $\lambda_{tv}$ as $5\times10^2$, $9\times10^3$,  $150$, and $2\times10^{-3}$, respectively. 
For the content loss, similar to Gatys et al.\cite{gatys}, we use the features of layers ``conv4\_2" and ``conv5\_2" for the content loss.

For the neural network $f$, we use lightweight U-net\cite{unet} architecture which has three downsample and three upsample layers, in which the channel sizes are 16,32 and 64 for each downsample layers. For stable training, we include sigmoid function at the last layer of $f$, so that the pixel value range is within the range of $[0,1]$. For training the network, we use Adam optimizer with learning rate of $5\times10^{-4}$. The total training iteration is set as 200, and we decreased the learning rate to half at the iteration of 100. {The training time is  about 40 seconds per text on a single RTX2080Ti GPU.}

For patch cropping, we use the patch size of 128 as our default setting since it shows best perceptual quality. We can obtain various effects with varying the crop size. The total number of cropped patches is set as $n=64$. 
For perspective augmentation, we use the function provided by Pytorch library. Detailed implementation is in Supplementary Materials.
For threshold rejection, we set $\tau$ as 0.7 in which the result has the best visual quality. 

In order to reduce the noise of text embedding, we use a prompt engineering technique  proposed by Radford et al.\cite{clip}. Specifically,
we make several texts with same meaning, and feed them to the text encoder. Then we use the averaged embedding instead of original single text conditions. {Finally, for better visualization for readers,  we apply same contrast enhancement techniques for all outputs including baselines.} Please refer to our Github repository: \href{https://github.com/cyclomon/CLIPstyler}{https://github.com/cyclomon/CLIPstyler}.

\begin{figure*}[t]
\vspace*{-0.5cm}
\centering
\includegraphics[width=1.0\linewidth]{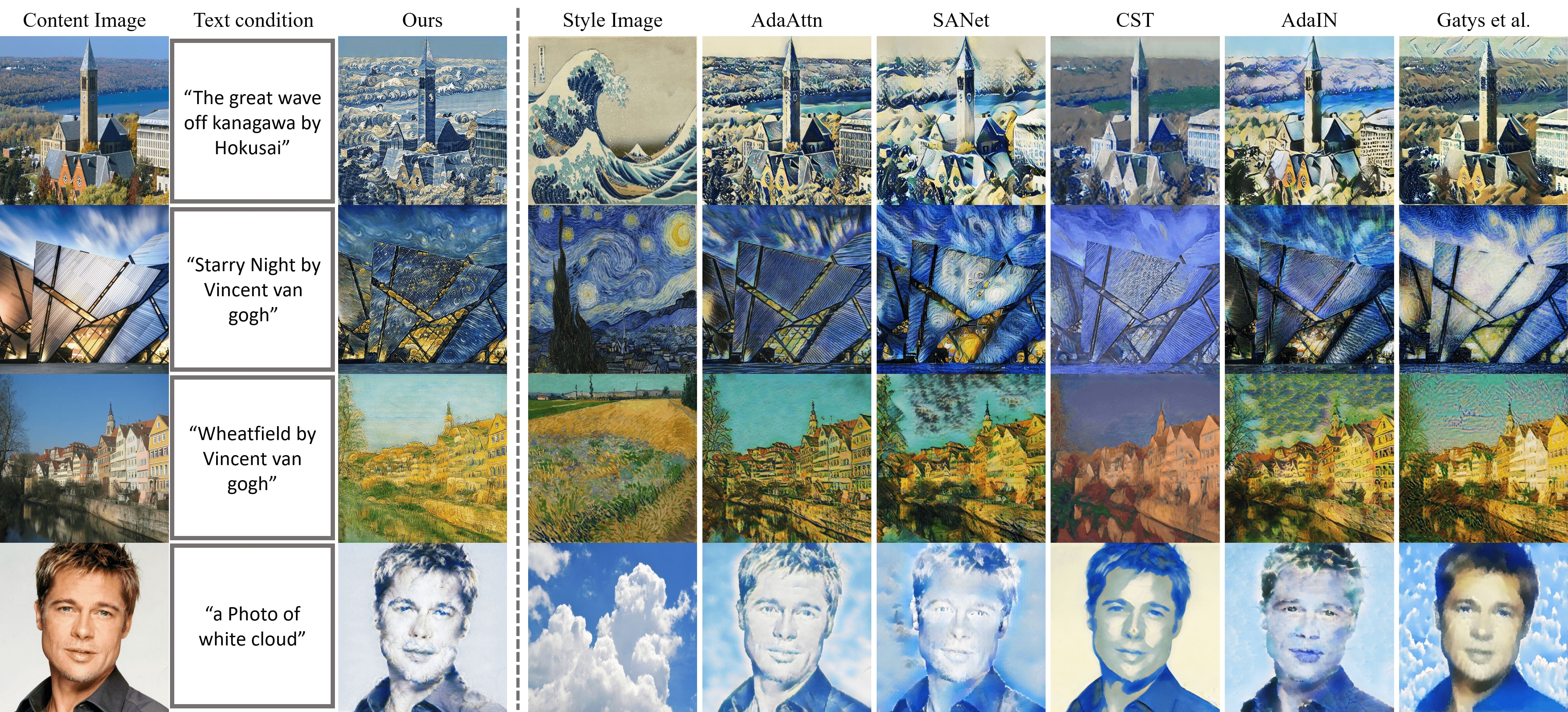}
\vspace*{-0.4cm}
\caption{Comparison results with baseline style transfer methods.  In contrast to the baselines, the results from our method is stylized only using a text condition without any style image. Although our method does not have style reference, the output image has complex texture which follows the semantics of the text conditions.
}
\label{fig:comp}
\end{figure*}

\subsection{Qualitative evaluations}
Figure~\ref{fig:first} and \ref{fig:ours} show representative style transfer results from our methods. With corresponding text conditions, we can successfully
convert image style that matches the text condition without changing the content of the image. 
In the result images, we can do style transfer not only for artistic styles, but also for a wide range of general texture styles. In particular, since we give conditions through text conditions, it is possible to control not only the texture type, but also the detailed conditions of each style. For example, we can give additional information of color with textures in using text conditions such as ``white wool" and ``green crystal". We can also choose which kind of `object' to use as target texture. In the results of Figure~\ref{fig:first}, we can choose the texture type (``oil painting") and pattern object (``flowers") simultaneously, so that more abundant effects can be obtained by our method.

\subsection{Comparison with baselines}
\noindent\textbf{Comparison with existing style transfer:}
Although our method does not follow the conventional framework of style transfer which require style images, we can indirectly compare our results with existing style transfer methods. Since the CLIP model is trained on a wide range of natural images as well as artistic images, we can transfer the styles of famous artworks with corresponding text conditions and compare them with the existing methods. For baselines, we choose various state-of-the-art artistic style transfer methods which includes arbitrary style transfer (AdaAttn\cite{liu}, SANet\cite{park}, CST\cite{svoboda}, and AdaIN\cite{adain}), and pixel optimization\cite{gatys}.

In Fig. \ref{fig:comp}, the results from our method show similar style transfer results to the baseline method in spite of using only text conditions. In the first to third rows, we compare the artistic style transfer performance of our text-guided style transfer with those of the baselines. For the case of early baseline models of AdaIN\cite{adain} and Gatys et. al.\cite{gatys}, the outputs  show limitation in expressing the target texture style with only concentrating on major color changes. The results by the recent style transfer methods of AdaAttn\cite{liu}, SANet\cite{park}, and CST\cite{svoboda}
have complex texture information with preserving the structure of content images. Our result has also vivid texture patterns of artworks in various locations, without damaging the structure of original content shape. 

 For further comparison with non-artistic styles, we include style transfer result in the last row of Fig. \ref{fig:comp}. Since the baseline style transfer methods are mostly trained on artworks, we can see that the baseline style transfer methods have difficulty in transferring the texture of the non-artistic styles. However, our method can successfully extract the semantic textures from the query text and apply them to the content image.

\begin{figure}[!t]
\centering
    \vspace*{-0.5cm}
  \includegraphics[width=1.0\linewidth]{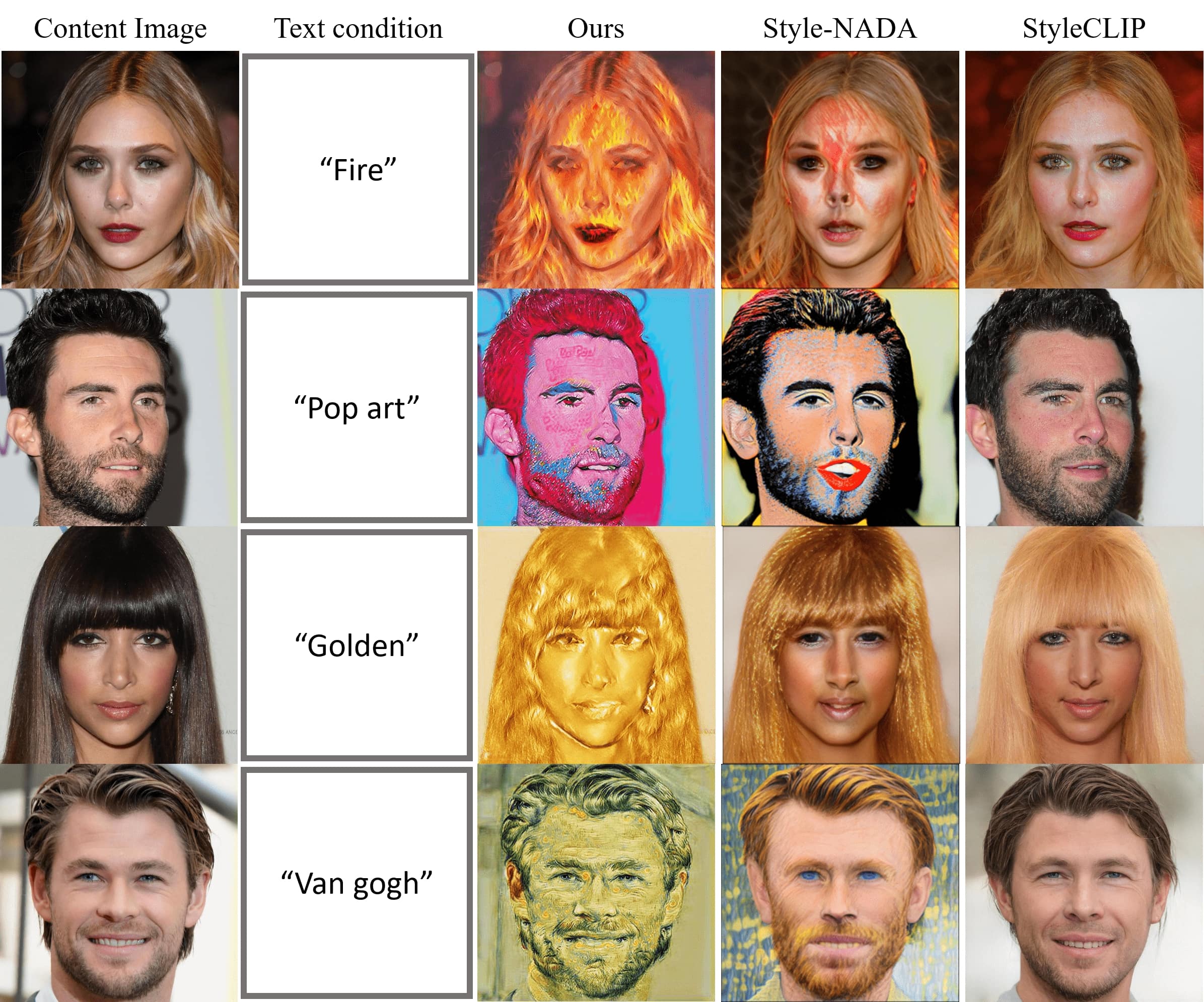}
  \vspace*{-0.5cm}
\caption{Comparison with other text-guided manipulation models. Our results have more realistic textures on the entire location. Baseline models show limited change, and in some cases the contents of the image are  modified.}
\label{fig:comp_text}
\end{figure}

\noindent\textbf{Comparison with text-guided manipulation models:} 
Figure~\ref{fig:comp_text} shows the result of comparison between our model and the baseline text-guided manipulation methods. As for baselines, we choose state-of-the-art methods which use CLIP model and a pre-trained StyleGAN. Since the baseline models are trained on human face, we experimented with images of CelebA-HQ\cite{celeba}. In the results, our method can express realistic textures on the content images, which are matched to the query text conditions. 

\begin{figure*}[!t]

\centering
\includegraphics[width=1.0\linewidth]{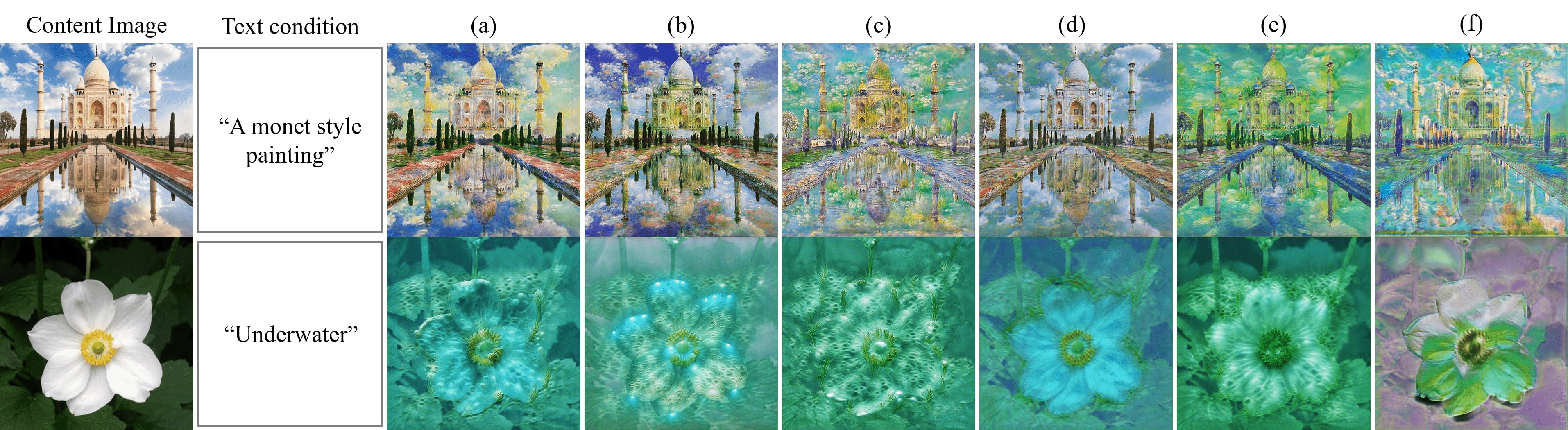}
\vspace*{-0.5cm}
\caption{Ablation study results. Each columns annotated with alphabets are the style transfer outputs (a) with using entire losses, (b) with removing $L_{dir}$, (c) with removing threshold rejection, (d) with removing augmentation, (e) with replacing perspective augmentation to random affine transformation, (f) with removing our proposed $L_{patch}$.
}
\label{fig:ablation}
\end{figure*}

\begin{figure}[!t]

\centering
\vspace*{-0.5cm}
\includegraphics[width=1.0\linewidth]{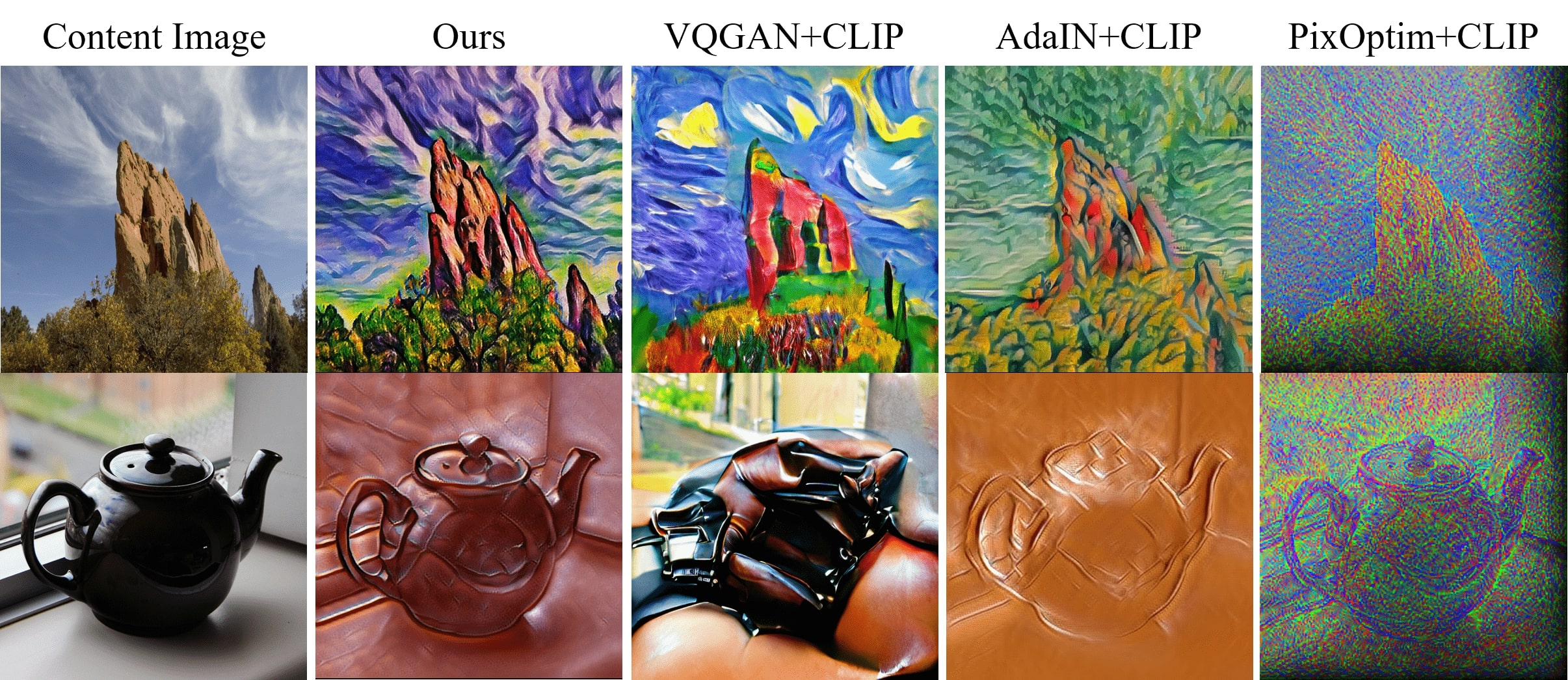}
\vspace*{-0.5cm}
\caption{Comparison study on various manipulation approaches. Top: outputs with ``A fauvism style painting". Bottom: outputs with ``Leather".
}
\vspace*{-0.5cm}
\label{fig:ablation_m}
\end{figure}

Specifically, in the result of StyleGAN-NADA\cite{nada}, the content images are modified to follow the text condition, but  only a part of the content image can be changed, or the changes do not sufficiently reflect the semantics of the text condition. The result in the last row of Figure~\ref{fig:comp_text} shows how our approach and StyleGAN-NADA differ in the way they accept text conditions. Since our model focus on textures, the model apply artistic style (e.g. Van Gogh style) to the entire image, but Style-NADA changed the face identity as it focused on the text condition itself.
In the results of StyleCLIP\cite{styleclip}, since the method can manipulate images within the learned latent domain, all of the results except for the third row failed to transform the images. 

For further comparison, as our method is based on the network weight optimization (or fine-tuning) using CLIP loss, we also investigate whether other manipulation approaches can produce better style transfer by combining with CLIP loss.
First, we combine CLIP-loss with the pixel optimization similar to Gatys et al.\cite{gatys}. Second, we only apply AdaIN code optimization using CLIP-loss on a pre-trained style transfer network  by Huang et al.\cite{adain}. Finally, we also compare with the latent code optimization on a pre-trained VQGAN\cite{vqgan} using CLIP-loss by
using the source code of VQGAN-CLIP\footnote{https://github.com/nerdyrodent/VQGAN-CLIP}. The results in Fig.~\ref{fig:ablation_m} show that our method of optimizing network $f$ shows superior quality compared to baselines. More specifically, with pixel optimization, the image could not reflect the semantic of text conditions. In case of VQGAN+CLIP and AdaIN+CLIP, the textures are applied to the contents, but the content structures are severely deteriorated. {In Supplementary Materials, we additionally show comparison results with two different baselines of 1) applying the existing style transfer method by using the text-
retrieved image as a style image, and 2) using
the image generated by the text-to-image generation model
as a style image.}
More comparisons with quantitative user study are also included in our Supplementary Materials.

\subsection{Ablation studies}
In order to verify the necessity of each component in our method, we conducted ablation studies. Figure \ref{fig:ablation} shows our ablation study results. When we use all the proposed loss functions (Fig.~\ref{fig:ablation}(a)), we can obtain the best results in perceptual domain. In particular, as well as three-dimensional texture, we could obtain a clean image without artifacts in terms of color. If the whole image CLIP loss $L_{dir}$ is not used (Fig.~\ref{fig:ablation}(b)), the global semantic cannot be captured, so we can see the color mapped in irregular patterns.  When the threshold rejection is removed (Fig.~\ref{fig:ablation}(c)), the image is over-focused on a specific patch and an over-stylized image is derived. When augmentation was not used (Fig.~\ref{fig:ablation}(d)),  the three-dimensional realistic texture was not reflected.  When using the commonly used affine transform instead of our proposed perspective augmentation (Fig.~\ref{fig:ablation}(e)), there are unwanted artifacts. Finally, when only directional CLIP Loss which is a loss function for the whole image was used, there is only little change in texture except for color (Fig.~\ref{fig:ablation}(f)).  {For further evaluation, we show user study results on various ablation experiments in Supplementary Materials. }

\section{Further Extensions}

\begin{figure}[!t]
\centering
    \vspace*{-0.5cm}
  \includegraphics[width=1.0\linewidth]{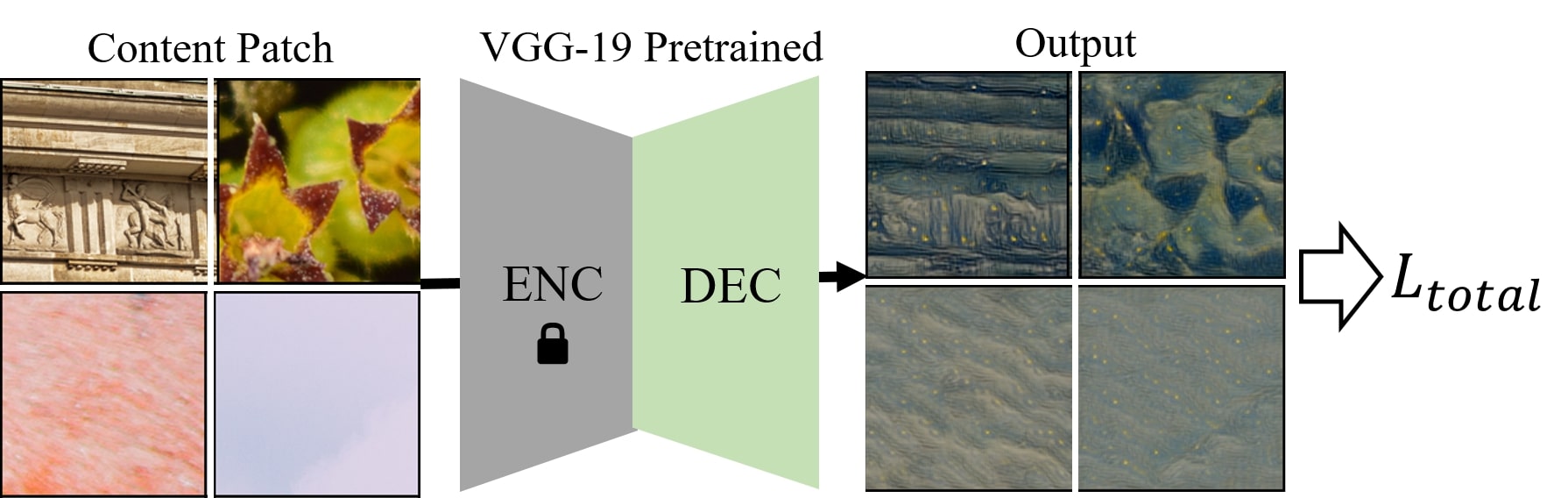}
  \vspace*{-0.5cm}
\caption{Training framework of our fast style transfer methods. We use cropped patches as content images, and used the same proposed loss functions to train the decoder of VGG19 networks.}
\vspace*{-0.5cm}
\label{fig:method_mult}
\end{figure}

\subsection{Fast style transfer}
In our default framework, we should train the style network $f$ for single content image to apply a given style. To overcome this, we are interested in a method that trains $f$ using various texture patches instead of single content image. Once the network is trained in such a way, the trained network can be used in various content images. 

Our proposed fast training scheme is illustrated in Fig.~\ref{fig:method_mult}. As a training set, we randomly cropped the patches from high-resolution texture images of DIV2k\cite{div2k}. For faster training, we employ   a pre-trained VGG encoder-decoder network rather than the U-Net as the style network $f$, and only the decoder network is fine-tuned. In training step, we used the same loss functions, but we did not crop the sub-patches because the input patches are already cropped from large images. We update the model for 200 steps with Adam optimizer using learning rate of $1\times10^{-4}$. The overall training time is $\sim$40 seconds. For inference, it take less than 0.5sec per content image. Further details are in  Supplementary Material.

Fig.~\ref{fig:result_mult} is the results from our fast style transfer method. Since we trained the model with diverse texture inputs, we can conduct style transfer on arbitrary content images in real-time.  We can see that the result images reflected the semantic texture of the text condition. Comparing with our results trained with single content image in Figs. \ref{fig:first} and \ref{fig:ours}, we can obtain similar texture transfer quality in our faster style transfer settings. Also, the results can adapt to any kind of content inputs with different structural diversities. However, in some cases (e.g. ``Neon Light",``Fire"), the images have textures on unnecessary regions such as background.  Still, the results show that our fast transfer enables high-quality style transfer for arbitrary content images.

\begin{figure}[!t]
\centering
  \includegraphics[width=1.\linewidth]{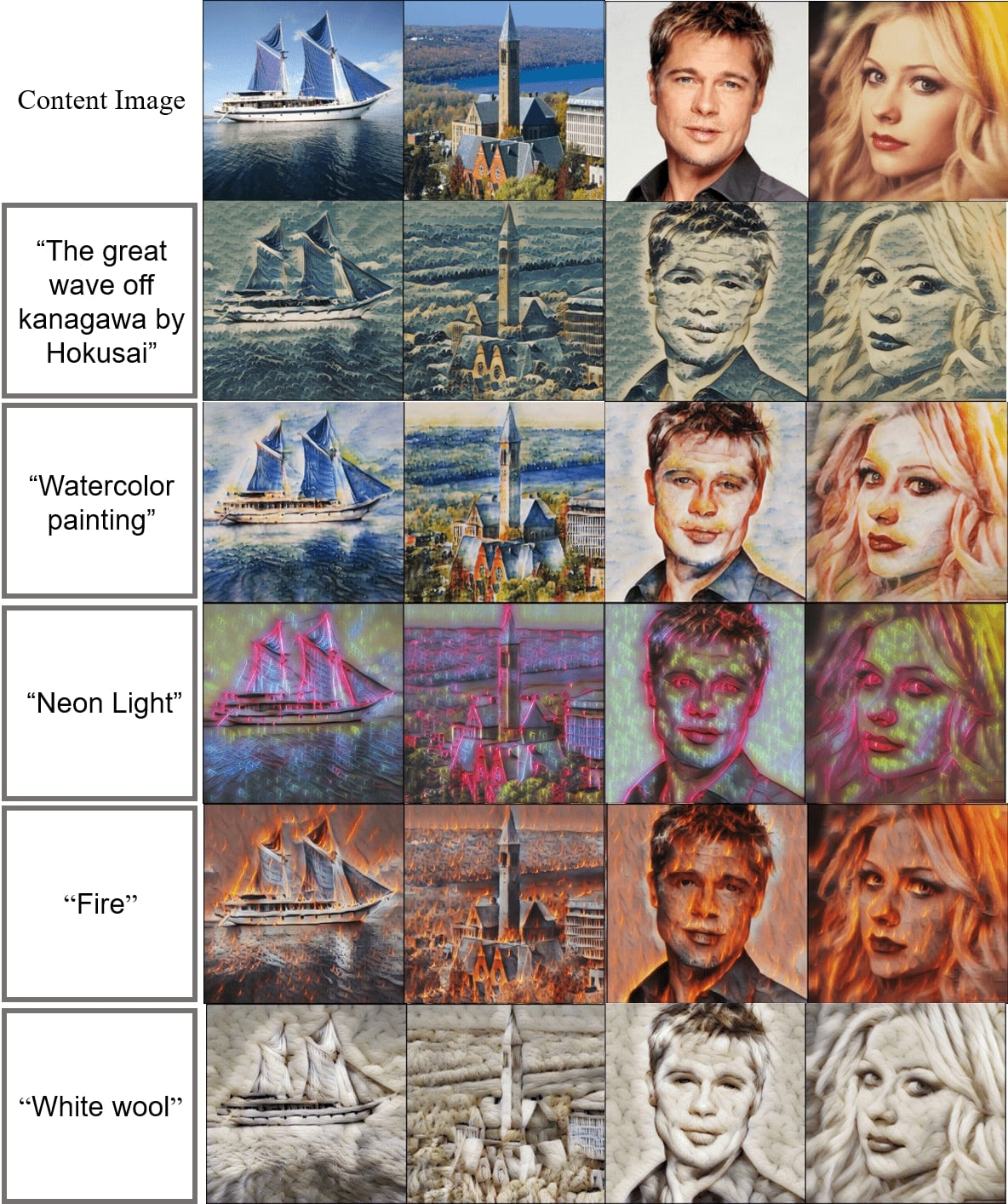}
  \vspace*{-0.5cm}
\caption{Results from our fast style transfer methods.}
\vspace*{-.4cm}
\label{fig:result_mult}
\end{figure}

\subsection{High-resolution style transfer}

\begin{figure}[!t]
\centering
    
  \includegraphics[width=1.0\linewidth]{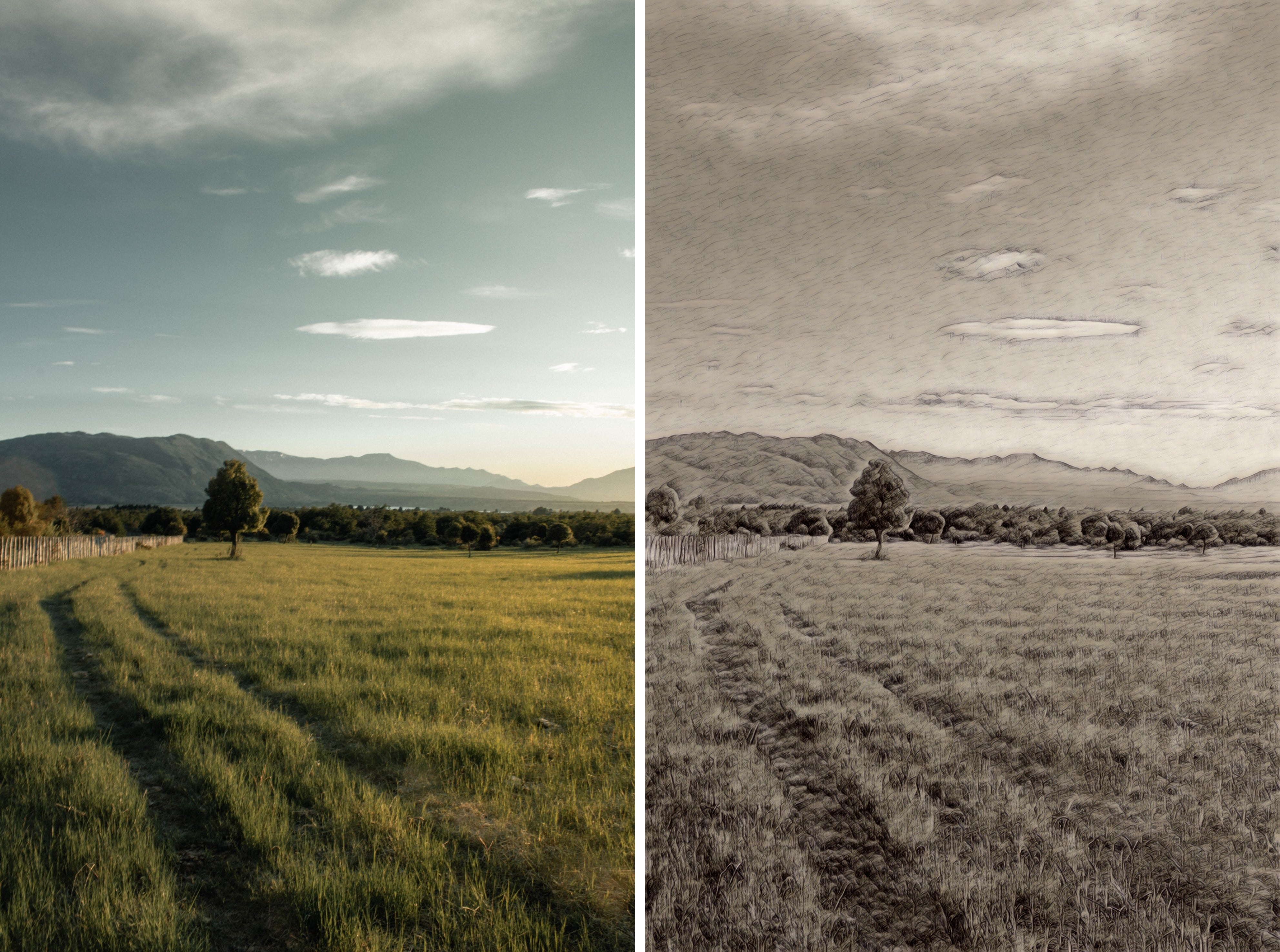}
\vspace*{-.5cm}
\caption{High-resolution style transfer results from our fast style transfer methods. Left: content image. Right: style transfer output with text condition of ``A sketch with black pencil". Image resolution is 3000$\times$2000. It takes about 4 seconds to process the image. }
 \vspace*{-.4cm}
\label{fig:highres}
\end{figure}

Since our fast style transfer can be adapted to any kinds of content images with patch-based training, we can transfer the style with higher resolution content images. The training scheme is same as our fast style transfer method. After training the VGG model, we can feed the high-resolution image to trained network for style transfer. In Figure~\ref{fig:highres}, we can change the style of input to given text conditions while maintaining the details of the content. More results are in Supplementary Materials.

\section{Conclusion}

In this paper, we proposed a novel image style transfer framework to transfer the semantic texture information only using text condition. 
Using novel  patchCLIP loss and augmentation scheme, we obtained realistic style transfer results by simply changing the text conditions without
ever requiring any style images. 
Experimental results demonstrated that our framework produced the state-of-the-art image style transfer. {For the discussions on limitations and social impacts, please refer to our Supplementary Materials.}
\\
\\
\noindent\textbf{Acknowledgement:} This research was supported by Field-oriented Technology Development Project for Customs Administration through National Research Foundation of Korea(NRF) funded by the Ministry of Science \& ICT and Korea Customs Service(NRF-2021M3I1A1097938), and supported by Institute of Information \& communications Technology Planning \& Evaluation (IITP) grant funded by the Korea government(MSIT) 
(No.2019-0-00075, Artificial Intelligence Graduate School Program(KAIST))


\begin{figure*}
\centering
\includegraphics[width=1.0\linewidth]{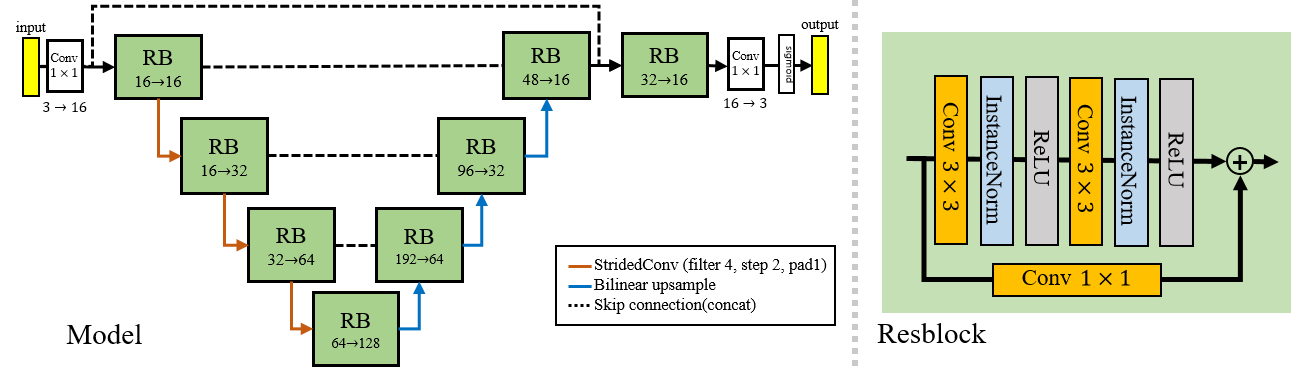}
 \vspace*{-0.2cm}
\captionof{figure}{Our style network model architecture.}
\label{fig:model}
\end{figure*}


\appendix


\section*{\Large{\textbf{Supplementary Material}}}
\section{More details of implementation}
\noindent\textbf{Network architecture}: For our style network $f$, we use a lightweight U-net which is described in the Fig.~\ref{fig:model}. We discovered that using residual block improves the content preservation and training stability. Since we train the network with only 200 iterations and the resolution of input image is relatively high, our network have to be lightweight. Therefore, we limited the maximum channel numbers to 128, and the highest resolution layer has only 16 channels. To better preserve the content information, we included another skip connection between input-output features. 

\noindent\textbf{Implementation details}: Since CLIP model receives the images with resolution of 224$\times$224, we resized all of the images including patch and whole image before feeding to the CLIP model. For augmentations, we use perspective augmentation which is implemented on Pytorch torchvision library. We directly used \verb|RandomPerspective(distortion_scale=0.5)|. The random perspective function is implemented in \verb|torchvision.transforms|.

\noindent\textbf{Details for fast style transfer}: For training our fast style transfer model, we used the cropped patches from DIV2K\cite{div2k}, and the crop size is 224. We used batch size of 4 and Adam optimizer with learning rate of $1\times10^{-4}$. Similar to our basic method, we used learning rate decay strategy. For augmentation, we applied random perspective augmentation for 16 times, therefore our total training patch number per iteration is $N=64$. We also calculated directional CLIP loss $L_{dir}$ with using patches before applying augmentations. Therefore, the loss functions of our fast style transfer is defined as:
\begin{equation*}
    L_{total} = \lambda_d L_{dir}+ \lambda_p L_{patch} + \lambda_c L_c + \lambda_{tv} L_{tv},
\end{equation*}
which is same as the loss of our basic method.
For hyperparameters, we set $\lambda_d$, $\lambda_p$, $\lambda_c$, and $\lambda_{tv}$ as $1$, $10$,  $1$, and $1\times10^{-4}$, respectively. For threshold rejection, we set $\tau$ as 0.7.  

For detailed implementation, please refer to our source code\footnote{https://github.com/paper11667/CLIPstyler}.

\section{Additional comparison results}

\begin{figure*}[!t]
\centering
\vspace{-0.3cm}
\includegraphics[width=0.9\linewidth]{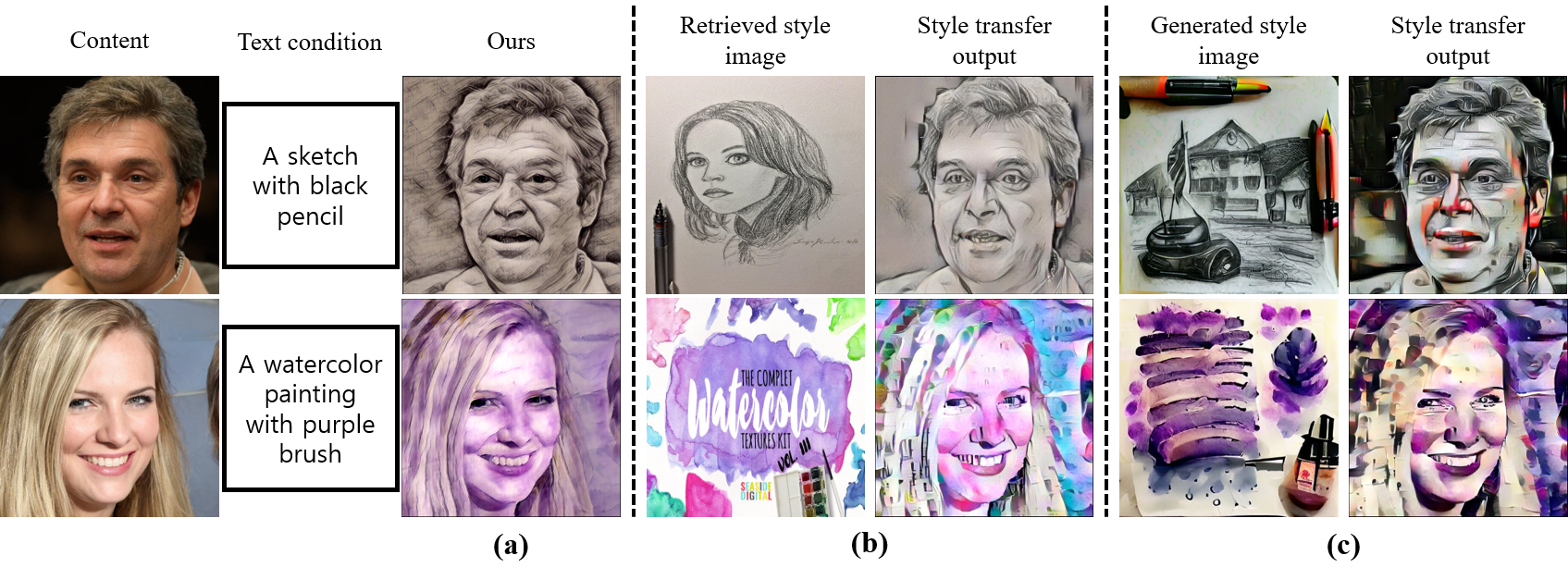}
\vspace{-0.5cm}
\caption{(a) Our results. (b) Retrieved style image and style transfer output. (c) Generated style image and style transfer output.}
\vspace{-0.5cm}
\label{fig:reb1}
\end{figure*}

\subsection{Comparison with other baselines}
  As a further comparison with other baselines, we show additional comparison results with two simple approaches: \textcolor{orange}{1)} to apply the existing style transfer method by using the text-retrieved image as a style image, and \textcolor{orange}{2)} using the image generated by the text-to-image generation model as a style image. For \textcolor{orange}{1)}, we used the pre-trained text-to-image retrieval model\footnote{https://github.com/rom1504/clip-retrieval} and chose the image with the highest CLIP score as a style image. For \textcolor{orange}{2)}, we used the image generated through VQGAN+CLIP\footnote{https://github.com/nerdyrodent/VQGAN-CLIP} as a style image. For image style transfer method, we chose widely used transfer model SANet\cite{park} in both of \textcolor{orange}{1)} and \textcolor{orange}{2)}.

In Fig.~\ref{fig:reb1}, we can observe that the outputs of baseline \textcolor{orange}{1)} and \textcolor{orange}{2)} both show reasonable results in the upper row. However, if the retrieved or generated style image is not adequate for style image (bottom row), we can see that the outputs are degraded. Although the frameworks are efficient in inference time, the results show that \textcolor{orange}{1)} and \textcolor{orange}{2)} methods have major disadvantages in that they are highly dependent on the performance of retrieval and generation models.

\begin{table}[!t]
\begin{center}

\centering
		\begin{adjustbox}{width=0.35\textwidth}
		\begin{tabular}[b]{c|cc}
        \multirow{2}{*}{\textbf{Methods}}&\multicolumn{2}{c}{\textbf{User study}}\\
      & \textbf{Style $\uparrow$} & \textbf{Content $\uparrow$} \\ \hline
      Retrieve + Sty \textcolor{orange}{1)} & 3.01& 3.38\\
      Text2Image + Sty \textcolor{orange}{2)} & 2.80& 3.61 \\
      \hdashline
      StyleCLIP & 1.47&\textcolor{red}{\textbf{4.35}}\\
      StyleGAN-NADA & 2.66&3.65\\
      \hdashline
      AdaIN+CLIP &\textcolor{blue}{\textbf{3.32}}&2.97\\
      VQGAN+CLIP & 2.75&2.31\\
      \hdashline
      Ours & \textcolor{red}{\textbf{3.78}}& \textcolor{blue}{\textbf{4.18}} \\
    \hline
    \end{tabular}
    \end{adjustbox}

\end{center}
\vspace*{-.3cm}
\caption{User study results on various text-guided image manipulation models. The values marked with blue color refer to  the best scores, and the values with red color are the second best scores.}

\vspace*{-.3cm}
\label{table:user}
\end{table}

\subsection{User study}
\noindent\textbf{Experiment Details:} For quantitative comparison, we conducted a user study. For baseline models, we selected six different text-guided manipulation methods: StyleGAN-NADA\cite{nada}, StyleCLIP\cite{styleclip}, AdaIN+CLIP, VQGAN+CLIP, and the previous approaches \textcolor{orange}{1)}, \textcolor{orange}{2)}.  Since the conventional style transfer methods use style images, we did not carry out experiments using these methods, and just selected the CLIP-based methods for fair comparison. 

For user study, we generated 160 different stylized images with 10 different text conditions for each model (total 1,120 images). We used both of artistic style (e.g. ``A sketch with black pencil") and non-artistic style (e.g. ``Leather") text conditions.  Since StyleGAN-NADA and StyleCLIP provided the models pre-trained on human face dataset, we used randomly sampled human face images as content images for fair comparison. 

With generated images, we created 20 different questions. Specifically, to quantify the user preference, we asked  participants questions about content preservation and the expression of textures that match the text conditions. In order to collect the detailed opinions of users, we used a custom-made opinion scoring system using Google Form. More specifically, we provided a total of 25 users with stylized images and asked them to rate the level of content preservation and text customization.
 We set minimum score as 1, and the maximum scores is 5. For each question, users can choose the scores among 5 different options:  1-very low , 2-low, 3-middle, 4-high, 5-very high. The 25 different subjects come from the age group between 20s and 40s, who are randomly recruited online. Then we reported the average values.

\begin{figure*}[!t]

\centering
\includegraphics[width=0.9\linewidth]{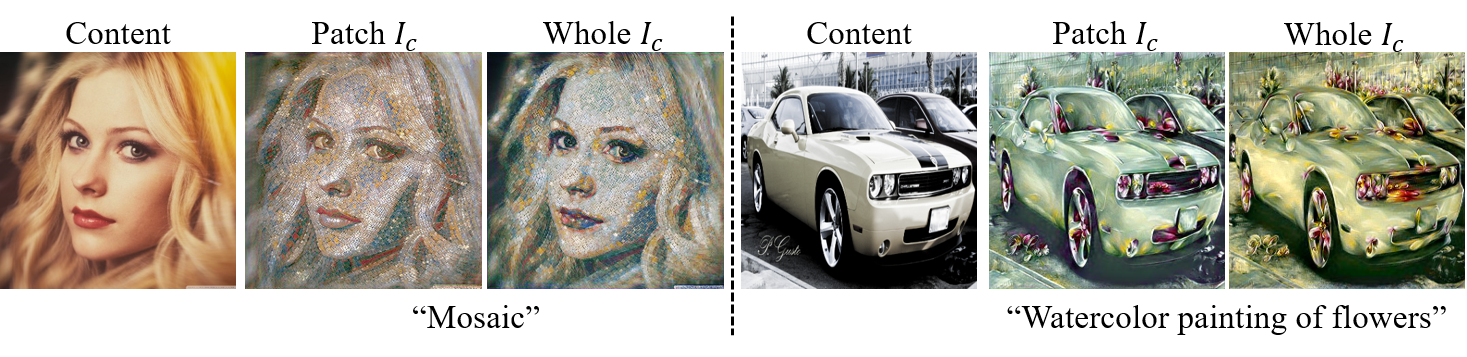}
\caption{Comparison result with patch cropping on content image $I_c$. The results show there is almost no difference in perceptual quality.}
\label{fig:content_crop}
\end{figure*}

\noindent\textbf{Results:} 
Table~\ref{table:user} shows the user study results on various text-guided models. Our model outperformed the baseline models in content and style scores. More specifically, StyleGAN-NADA showed decent score in content preservation, but the stylization score was relatively lower than other baselines. StyleCLIP showed the best score in content preservation, but it showed the worst score in stylization. StyleGAN-NADA and StyleCLIP showed bad results in style change as they are strongly confined to the pre-trained dataset using a pre-trained StyleGAN.  For baselines of \textcolor{orange}{1)} and \textcolor{orange}{2)}, the method showed decent scores in content preservation, but the methods failed to successfully transfer the target style with showing relatively low values in style scores.

In AdaIN+CLIP, the model shows much better style transfer performance with scoring the second best among all the models, but it has disadvantages in content preservation. For VQGAN+CLIP, the model shows degraded performance with the worst content preservation score, which means that the model hardly reflected the shape of the input contents. 

For our model, we obtained the second best score in content preservation with little difference to StyleCLIP, and obtained the best score in stylization. The user study results show that our model showed best performance considering both of content preservation and stylization.

\subsection{Patch-wise CLIP score}

\begin{table}[!t]
\vspace{-0.2cm}

		\centering
		\begin{adjustbox}{width=0.3\textwidth}
		\begin{tabular}[b]{c|c}
        \textbf{Methods}&\textbf{CLIP score $\uparrow$}\\
        \hline
      Retrieve + Sty \textcolor{orange}{1)} & 0.2317\\
      Text2Image + Sty \textcolor{orange}{2)} &0.2213 \\
      \hdashline
      StyleCLIP &  0.1982\\
      StyleGAN-NADA &0.2252\\
      \hdashline
      AdaIN+CLIP & \textcolor{blue}{\textbf{0.2487}}\\
      VQGAN+CLIP & 0.2249\\
      \hdashline
      Ours &\textcolor{red}{\textbf{0.2515}} \\
    \hline
    \end{tabular}
    \end{adjustbox}

	\captionof{table}{Quantitative comparison results on patch-wise CLIP scores. \textcolor{blue}{Blue-second best}, \textcolor{red}{Red-best}}
	\label{tab:clip}
\end{table}

In order to strengthen the evaluation, we measured additional quantitative metrics. To measure the correspondence between text condition and texture, we calculated cosine similarity between the CLIP embeddings of output patches and target texts. With single output image of 512$\times$512 resolution size, we randomly cropped 64 patches which have various resolution sizes (64$\times$64 $\sim$ 224$\times$224). As a validation set, we used the same images in the user study questions.

In Tab. \ref{tab:clip}, we show the averaged CLIP scores on various models.  Again, our model outperformed baseline models with showing highest CLIP scores. More specifically, StyleCLIP showed the worst performance with the lowest CLIP score, and AdaIN+CLIP scored second best among all of the baseline models. This shows that the CLIP score results have almost same tendencies as the stylization score of user study.

\section{Additional ablation studies}
\subsection{Patch crop on content images}
In order to figure out whether patch cropping on content image $I_c$ affects the output quality, we compared the outputs using proposed CLIP loss with cropped patches from both of content and output images. In Fig.~\ref{fig:content_crop}, we see little difference in perceptual quality, therefore we did not crop the patches from content images due to computation time. 
Note that  to obtain the loss with patches from the content image, more images should be embedded into the CLIP space, and it increases the run time ($\sim$10 seconds in our case) for each style transfer.

\begin{figure*}[!t]

\centering
\includegraphics[width=0.9\linewidth]{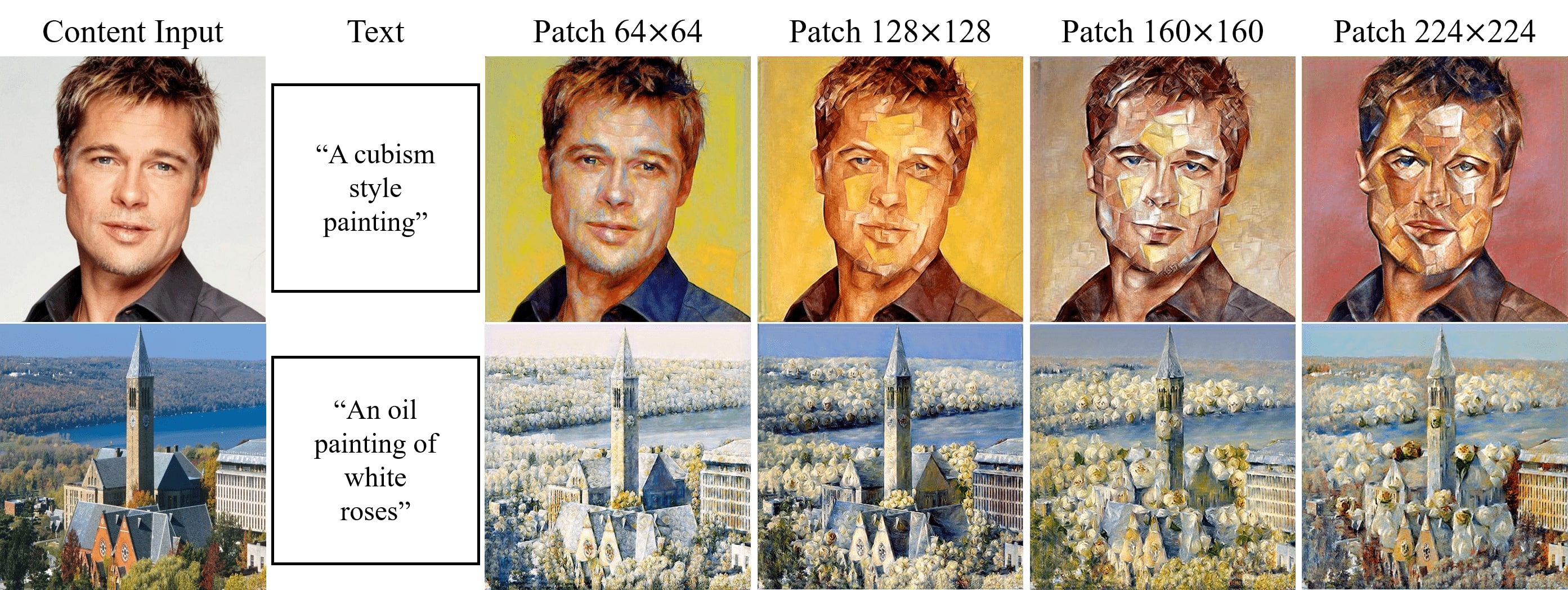}
\caption{CLIPstyler outputs with different patch crop sizes. With a smaller patch size, we can obtain finer stylization results. With a large patch size, the outputs have large style patterns. 
}
\label{fig:patch}
\end{figure*}

\subsection{User study}
For more thorough ablations study, we conducted another user study for ablation study in Tab. \ref{tab:abl}. For baselines, we use four different settings: training without $L_{dir}$, without proposed $L_{patch}$, without proposed threshold rejection, and without augmentations. For each setting, we generated 80 different stylized images with 10 text conditions (total 400 images). 

With generated images, we created 10 different questions. In order to collect the detailed opinions of users, we used a custom-made opinion scoring system using Google Form. More specifically, we provided a total of 20 users with stylized images and asked them to score the images with overall preference. Users can choose the scores among 5 options (1-very bad, 2-bad, 3-neutral, 4-good, 5-very good). The results show that the users prefer our best setting. The 20 different subjects come from the age group between 20s and 40s, who are randomly recruited online.

The results show that our best setting outperforms other baselines with obtaining higher preference scores. We can see that without using our proposed patch-wise CLIP loss (no $L_{patch}$), the images showed the worst score among all of the baselines, which means that $L_{patch}$ is the most important loss in stylization. 

\begin{table}[!t]
		\centering
		\begin{adjustbox}{width=0.3\textwidth}
		\begin{tabular}[b]{c|c}
      \textbf{Settings} & \textbf{Preference Score$\uparrow$} \\ \hline
      
      no Augment &2.61\\
      no Thresh & \textcolor{blue}{\textbf{3.02}} \\
      no $L_{patch}$ & 1.69 \\
      no $L_{dir}$ & 2.58\\
      \hdashline
      $L_{total}$ & \textcolor{red}{\textbf{3.92}} \\
      
    \hline
    \end{tabular}
    \end{adjustbox}
	
	\captionof{table}{User study for ablation study. \textcolor{blue}{Blue-second best}, \textcolor{red}{Red-best}}
	\label{tab:abl}

\end{table}

\section{Effect of patch size}

For calculating our proposed $L_{patch}$, we need to decide the proper patch crop size. Although we choose patch size of 128 as our default setting, we can obtain various effects with changing the patch size. The results in Fig.~\ref{fig:patch} shows the effect of different patch sizes. If we use larger patch size in training, we can have larger `brushstroke' which can stylize the content image in a rough scale.  With small patch sizes, we can apply much finer style patterns to the content images.

\section{Failure cases}

\begin{figure}[!t]

\centering
\includegraphics[width=1.0\linewidth]{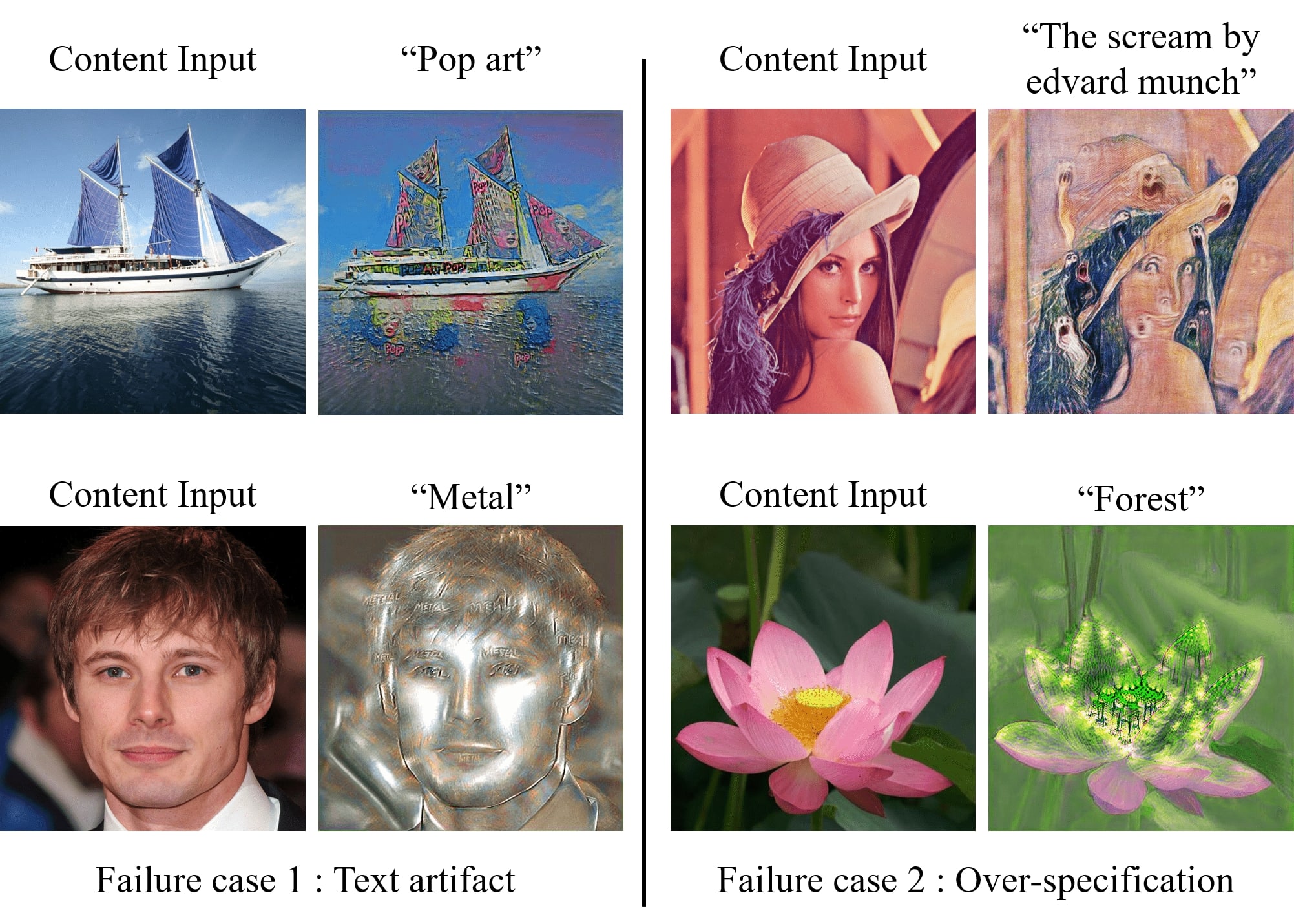}
\vspace*{-0.5cm}
\caption{Failure cases of our CLIPstyler.
}
\label{fig:failure}
\end{figure}

In this section, we show several failure cases of our model. First, as our model is based on random patch sampling and augmentations, failure cases often occur when bad patches are sampled. In the left panel of  Fig.~\ref{fig:failure},  we can see that there are artifacts which are caused by direct visualization of the text condition itself. For example, there are direct visualization of the text ``Pop art" on the first image. Although we found out that using slightly larger value of total variation loss can partially alleviate the artifact, it is not a perfect solution. In our future work, we plan to improve the sampling algorithm so that the sampled patches are perceptually meaningful in training the network. 

In the second case, output images with too specific text conditions may have unwanted patterns which are perceptually bad. In the right panel of Fig.~\ref{fig:failure}, the model overly focused on reconstructing the text-conditioned target `object', instead of applying the `texture' of text conditions. Since this artifact is mainly caused by the embedding aspect of pre-trained CLIP model, we need to detour the effect with using different text conditions of similar meanings, such as ``A painting of forest", instead of ``Forest".


\section{Additional results}
\begin{figure*}[p]
    \includegraphics[width=0.99\linewidth]{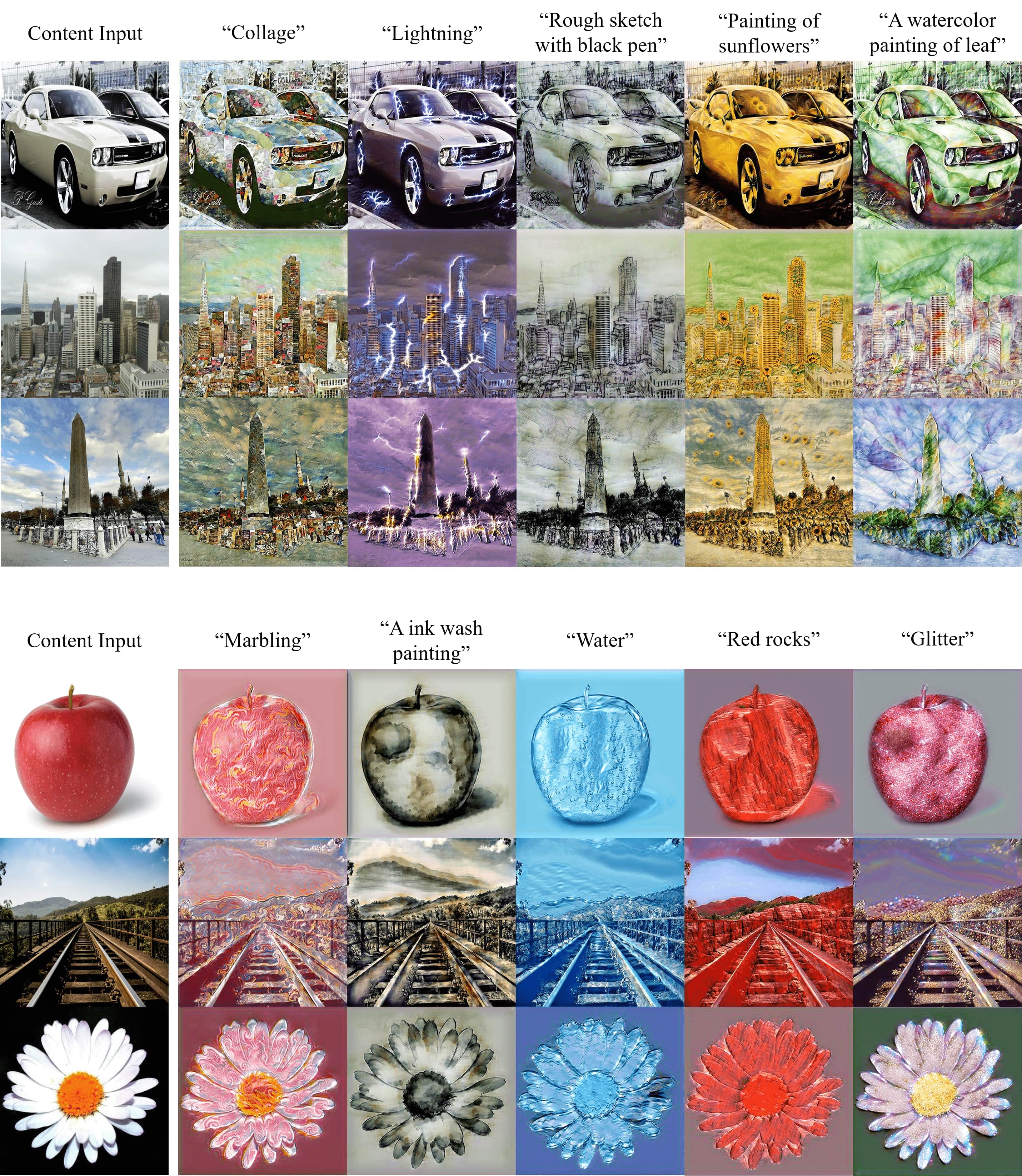}
    \centering
    \vspace*{-0.2cm}
    \caption{Additional results of our CLIPstyler.}
    \label{fig:addResult_mult}
    \vspace*{-0.4cm}
\end{figure*}

\begin{figure*}[p]
    \includegraphics[width=0.99\linewidth]{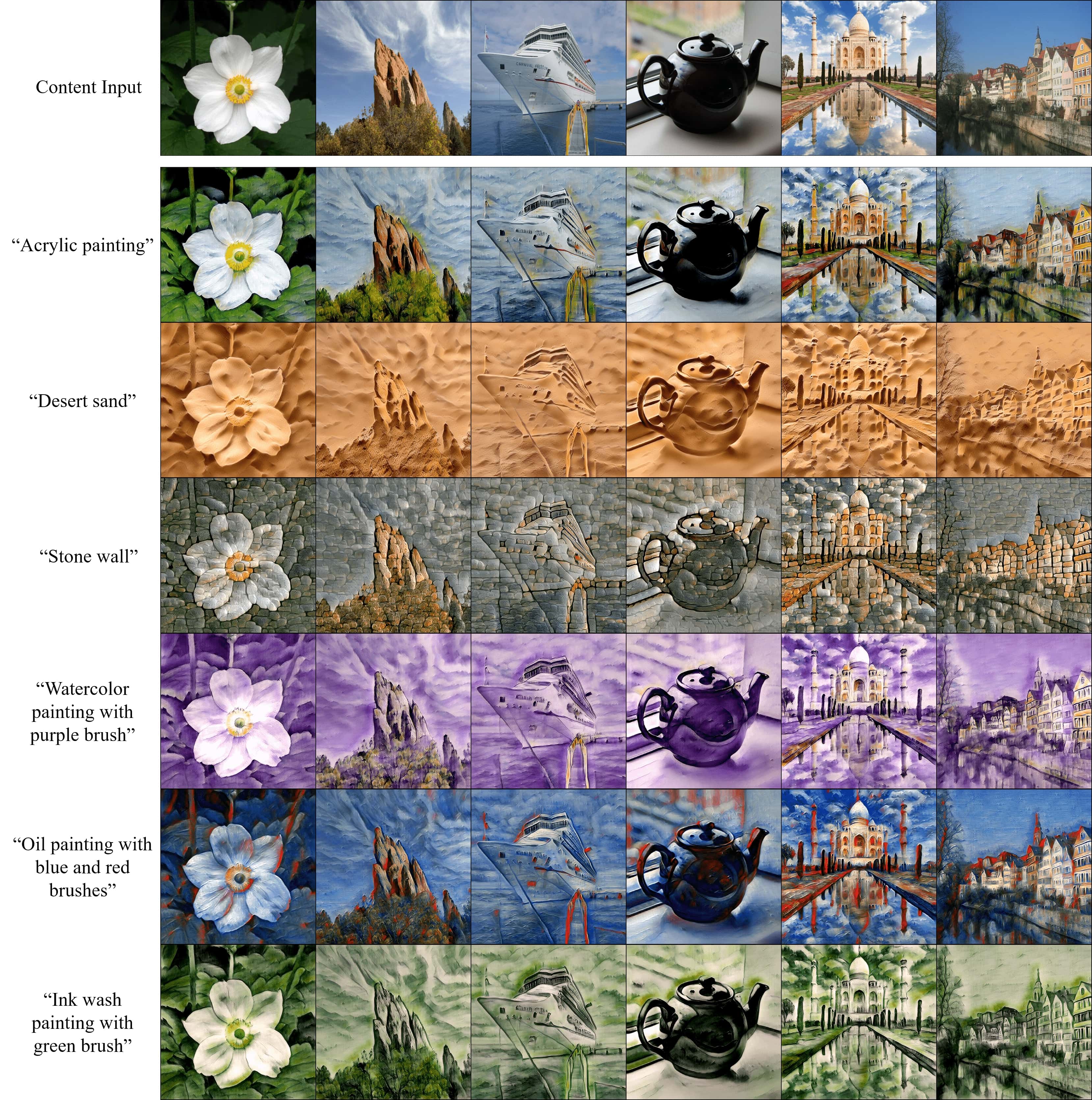}
    \centering
    \vspace*{-0.2cm}
    \caption{Additional results of our fast style transfer method.}
    \label{fig:addResult_fast}
    \vspace*{-0.4cm}
\end{figure*}

\noindent\textbf{More results of CLIPstyler:} In Fig.~\ref{fig:addResult_mult}, we show our text-guided style transfer outputs on various content images.
The results clearly demonstrate that our method can apply realistic textures to content images. We also show additional results on our fast style transfer method in Fig.~\ref{fig:addResult_fast}.

\begin{figure*}[p]
    \includegraphics[width=0.99\linewidth]{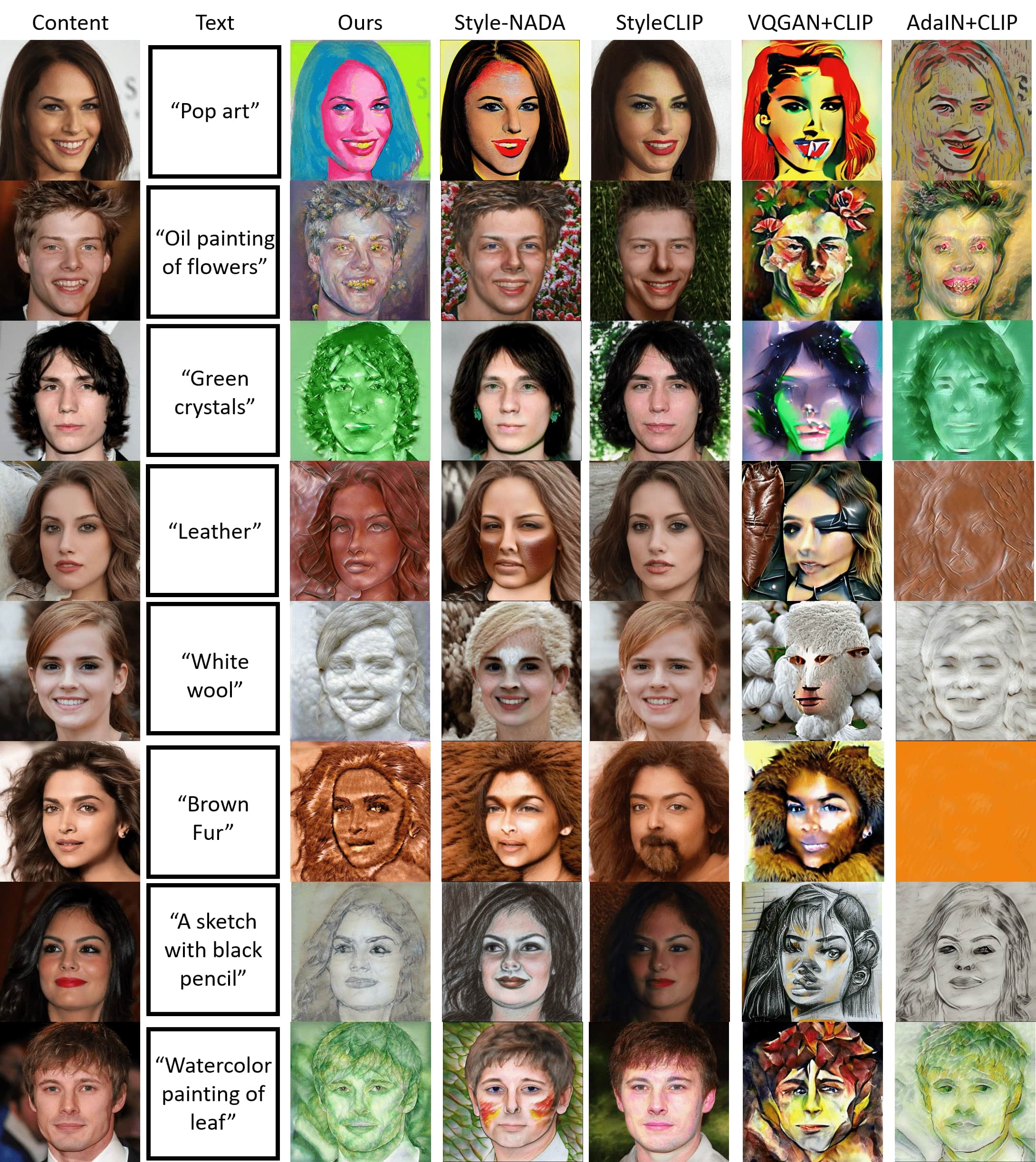}
    \centering
    \vspace*{-0.2cm}
    \caption{Additional comparison with baseline methods.}
    \label{fig:addResult_comp}
    \vspace*{-0.4cm}
\end{figure*}

\begin{figure*}[p]
    \includegraphics[width=0.99\linewidth]{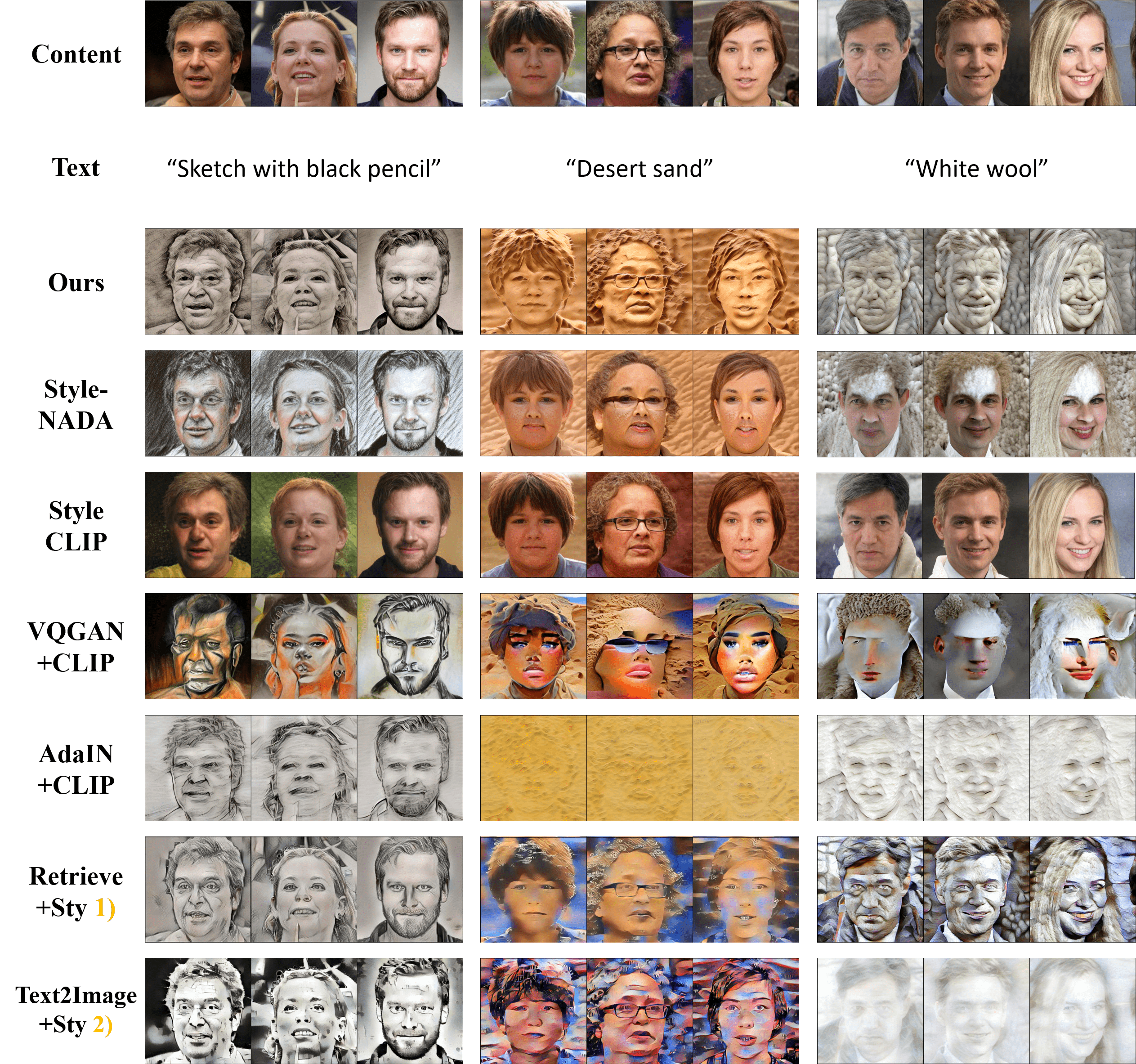}
    \centering
    \vspace*{-0.2cm}
    \caption{Additional comparison with baseline methods.}
    \label{fig:addResult_comp2}
    \vspace*{-0.4cm}
\end{figure*}

\noindent\textbf{More comparison results:} For further qualitative comparison, we provide style transfer outputs from various baseline models which use CLIP to manipulate the images. We compare the results from our model, Retrieve + Sty \textcolor{orange}{1)}, Text2Image + Sty \textcolor{orange}{2)}, StyleGAN-NADA, StyleCLIP, VQGAN+CLIP, and AdaIN+CLIP. Since StyleGAN-NADA and StyleCLIP are trained on human face dataset, we carried out experiments  on various human face images. Part of provided results are also used in our user study questions. 
In Fig. \ref{fig:addResult_comp} and \ref{fig:addResult_comp2}, the results from our model show the best style transfer quality in both of content preservation and text-guided texture synthesis.

\begin{figure*}[p]
    \includegraphics[width=0.99\linewidth]{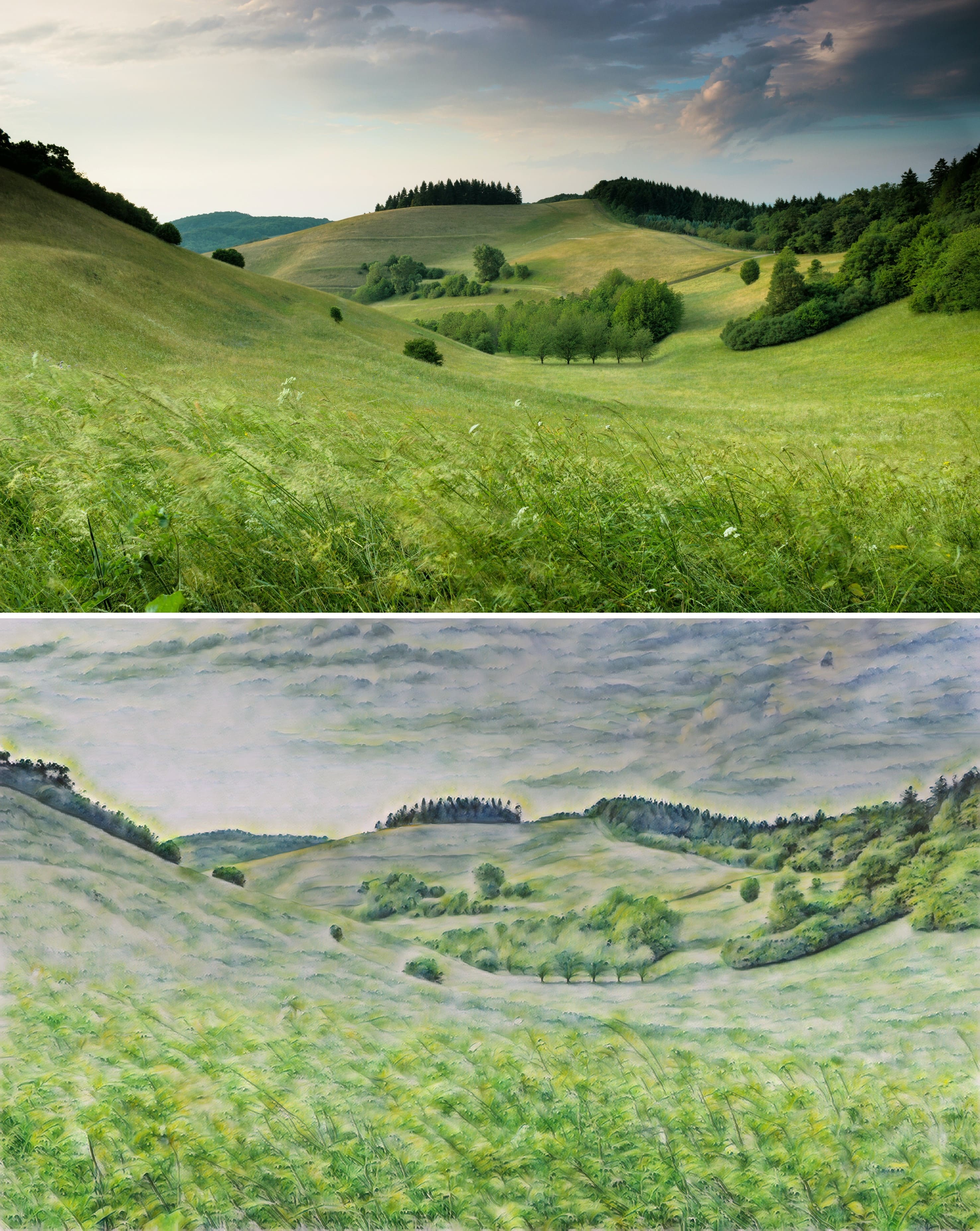}
    \centering
    \vspace*{-0.2cm}
    \caption{High resolution results from our fast style transfer method. The resolution size is 1920$\times$2580. Up: Content image. Down: Output with text condition of ``Watercolor painting".}
    \label{fig:addResult_high1}
    \vspace*{-0.4cm}
\end{figure*}

\begin{figure*}[p]
    \includegraphics[width=0.99\linewidth]{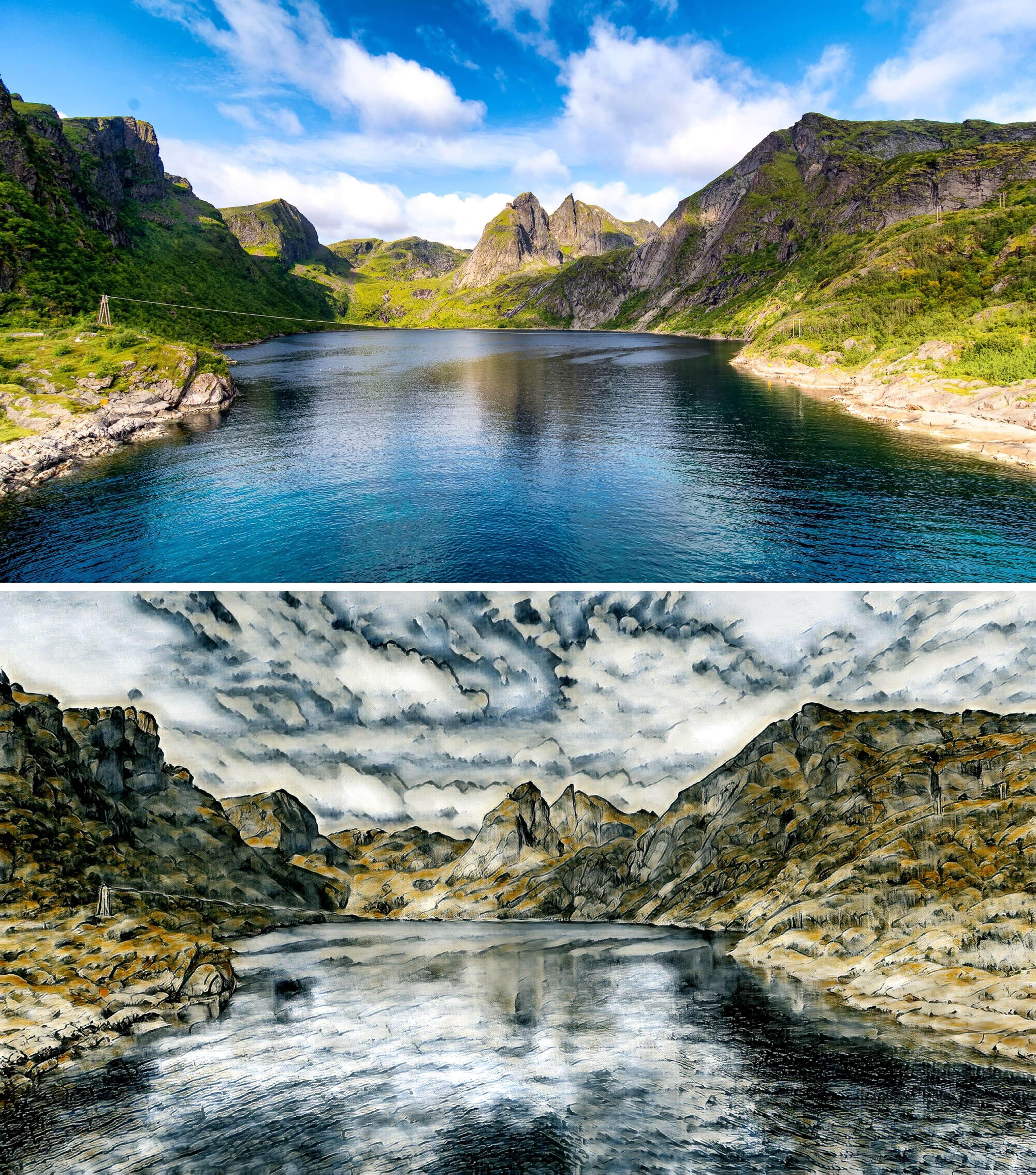}
    \centering
    \vspace*{-0.2cm}
    \caption{High resolution results from our fast style transfer method. The resolution size is 1128$\times$2000. Up: Content image. Down: Output with text condition of ``An ink wash painting".}
    \label{fig:addResult_high2}
    \vspace*{-0.4cm}
\end{figure*}

\begin{figure*}[p]
    \includegraphics[width=0.90\linewidth]{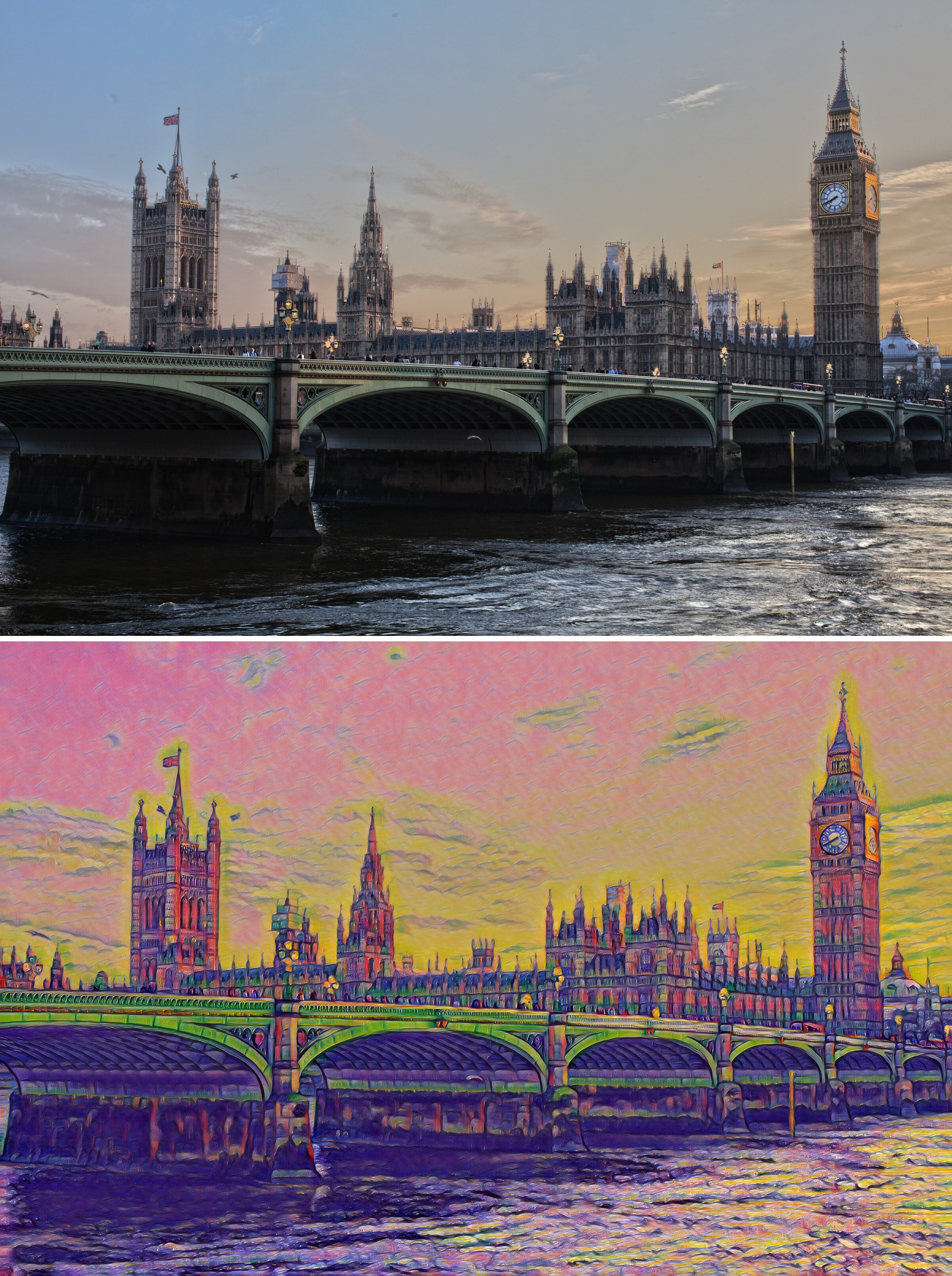}
    \centering
    \vspace*{-0.2cm}
    \caption{High resolution results from our fast style transfer method. The resolution size is 1600$\times$2400. Up: Content image. Down: Output with text condition of ``A fauvism style painting with bright color".}
    \label{fig:addResult_high3}
    \vspace*{-0.4cm}
\end{figure*}

\begin{figure*}[p]
    \includegraphics[width=0.85\linewidth]{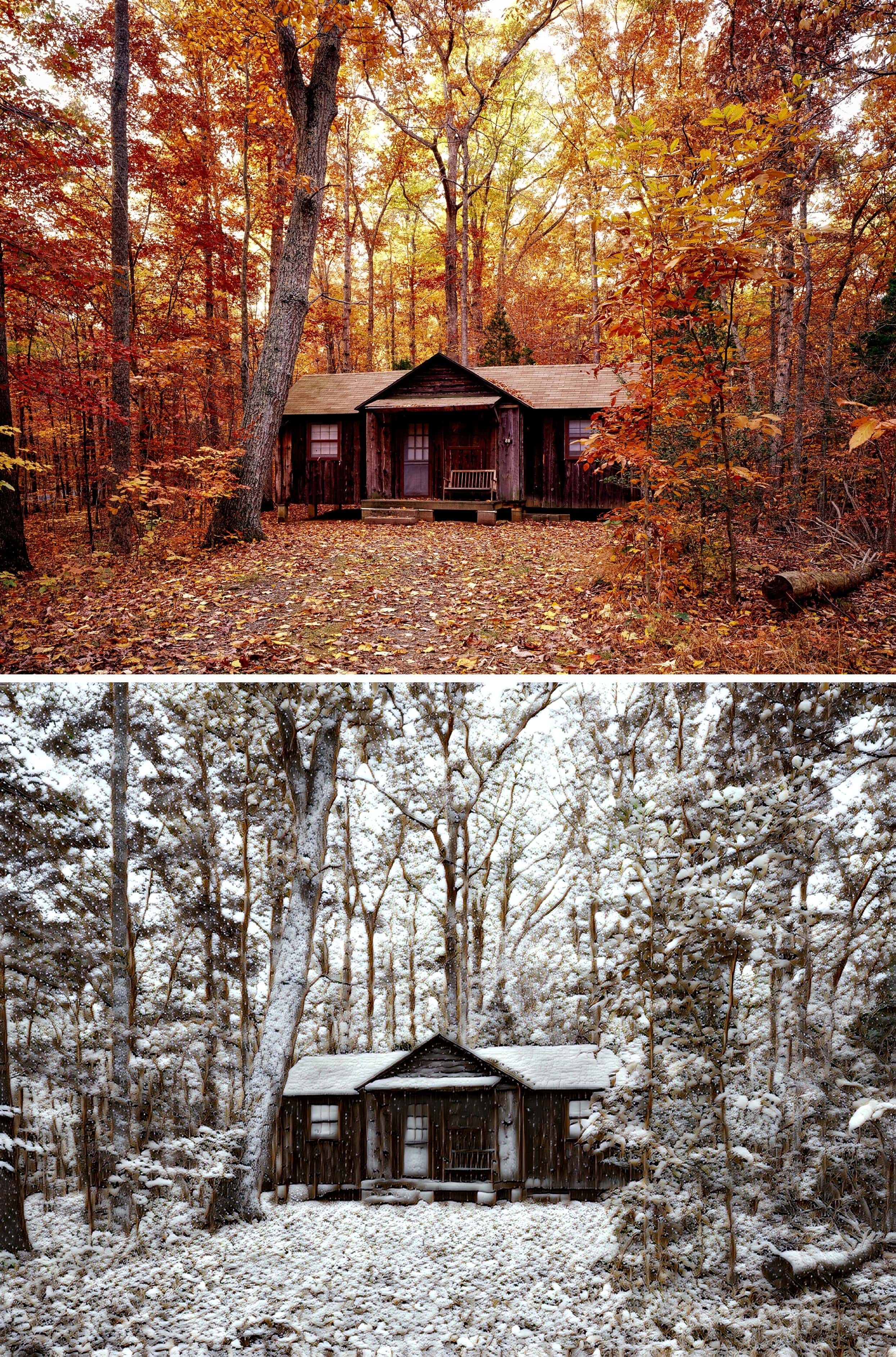}
    \centering
    \vspace*{-0.2cm}
    \caption{High resolution results from our fast style transfer method. The resolution size is 1656$\times$2200. Up: Content image. Down: Output with text condition of ``Snowy".}
    \label{fig:addResult_high4}
    \vspace*{-0.4cm}
\end{figure*}

\noindent\textbf{More high-resolution results:} We  also provide additional high-resolution output images from our fast style transfer method. The results in Figs.~\ref{fig:addResult_high1},\ref{fig:addResult_high2},\ref{fig:addResult_high3},\ref{fig:addResult_high4} shows the results. Our model can synthesize the realistic textures for high-resolution images from text conditions.

\section{Limitations and Negative social impact}
Although our method shows high-quality results with single text condition, there are several remaining technical issues. First,  our transfer method requires network training with each given text condition, so real-time style transfer  is still not possible. 
While our fast style transfer method can partially address this issue,  we observe that the resulting  quality of the style transfer is
still inferior to our default single image based optimization. Second, as shown in our failure cases,  low quality results can be produced when bad patches are sampled. As a future work, we will try to solve the problems explained.

Since our method manipulate the content images with various text conditions, when the user manipulate the content images with malicious words such as obscene expressions, it may bring negative social effects. 
In addition, when such malicious style transfer is applied to personal photos containing sensitive information, the impact may be greater.

{\small
\bibliographystyle{ieee_fullname}
\bibliography{egbib}
}

\end{document}